\documentclass{article}

\usepackage[final]{neurips_2021}

\usepackage[utf8]{inputenc} %
\usepackage[T1]{fontenc}    %
\usepackage{url}            %
\usepackage{booktabs}       %
\usepackage{amsfonts}       %
\usepackage{nicefrac}       %
\usepackage{microtype}      %
\usepackage{xcolor}         %

\usepackage{graphicx}
\usepackage{caption}
\usepackage{algorithm}
\usepackage{algorithmic}
\usepackage{wrapfig}

\usepackage{tabularx}
\usepackage{booktabs}
\usepackage{multirow}

\usepackage{subcaption}
\usepackage{amsmath}
\usepackage{amssymb}
\usepackage{xparse}

\newcommand{\mdpm}{\mathcal{M}}
\newcommand{\mdps}{\mathcal{S}}
\newcommand{\mdpa}{\mathcal{A}}

\newcommand{\mdpp}{\mathcal{P}}
\newcommand{\mdprho}{\rho_0}

\newcommand{\buffer}{\mathcal{B}}
\newcommand{\tasks}{\mathcal{T}}

\usepackage{color}

\NewDocumentCommand\bbm{}{ \begin{bmatrix} }
\NewDocumentCommand\ebm{}{ \end{bmatrix} }

\NewDocumentCommand\Real{}{ \mathbb{R} }

\usepackage{hyperref}
\usepackage{cleveref}

\title{Learning from Guided Play: A Scheduled Hierarchical Approach for Improving Exploration in Adversarial Imitation Learning}

\author{%
  Trevor Ablett$^*$, Bryan Chan$^*$, Jonathan Kelly \\
  University of Toronto, Canada\\
}

\begin{document}

\maketitle

\begin{abstract}

Effective exploration continues to be a significant challenge that prevents the deployment of reinforcement learning for many physical systems.
This is particularly true for systems with continuous and high-dimensional state and action spaces, such as robotic manipulators.
The challenge is accentuated in the sparse rewards setting, where the low-level state information required for the design of dense rewards is unavailable.
Adversarial imitation learning (AIL) can partially overcome this barrier by leveraging expert-generated demonstrations of optimal behaviour and providing, essentially, a replacement for dense reward information.
Unfortunately, the availability of expert demonstrations does not necessarily improve an agent's capability to explore effectively and, as we empirically show, can lead to inefficient or stagnated learning.
We present Learning from Guided Play (LfGP), a framework in which we leverage expert demonstrations of, in addition to a main task, multiple auxiliary tasks.
Subsequently, a hierarchical model is used to learn each task reward and policy through a modified AIL procedure, in which exploration of all tasks is enforced via a scheduler composing different tasks together.
This affords many benefits: learning efficiency is improved for main tasks with challenging bottleneck transitions, expert data becomes reusable between tasks, and transfer learning through the reuse of learned auxiliary task models becomes possible.
Our experimental results in a challenging multitask robotic manipulation domain indicate that our method compares favourably to supervised imitation learning and to a state-of-the-art AIL method.\footnote{Code available at \url{https://github.com/utiasSTARS/lfgp}.}
\end{abstract}

\section{Introduction} \label{sec:intro}

Exploration is a crucial part of effective reinforcement learning (RL).
A variety of methods have been developed for attempting to optimize the exploration-exploitation trade-off of RL agents \cite{sutton2018reinforcement, bellemareUnifyingCountBasedExploration2016b, nairOvercomingExplorationReinforcement2018}, but the development of a method that generalizes across domains is still very much an open research problem.
A simple, well-known approach for improving learning efficiency is to provide a dense, or ``shaped'' reward to learn from, but this can be very challenging to design appropriately \cite{ngShapingPolicySearch2003a}.
Furthermore, the environment may not directly provide the low-level state information required for such a reward.
An alternative to providing a dense reward is to learn a reward function from expert demonstrations of a task, in a process known as inverse RL (IRL).
Most current, state-of-the-art approaches to IRL are in the class known as adversarial imitation learning (AIL).
In AIL, rather than learning a true reward function, the policy and the discriminator form a two-player mini-max optimization problem: the policy aims to confuse the discriminator by minimizing the objective while the discriminator aims identify expert-like data by maximizing the objective \cite{ho2016generative, fu2018learning, kostrikov2019imitation}. %

\begin{figure*}[ht]
	\centering
	\includegraphics[width=.95\linewidth]{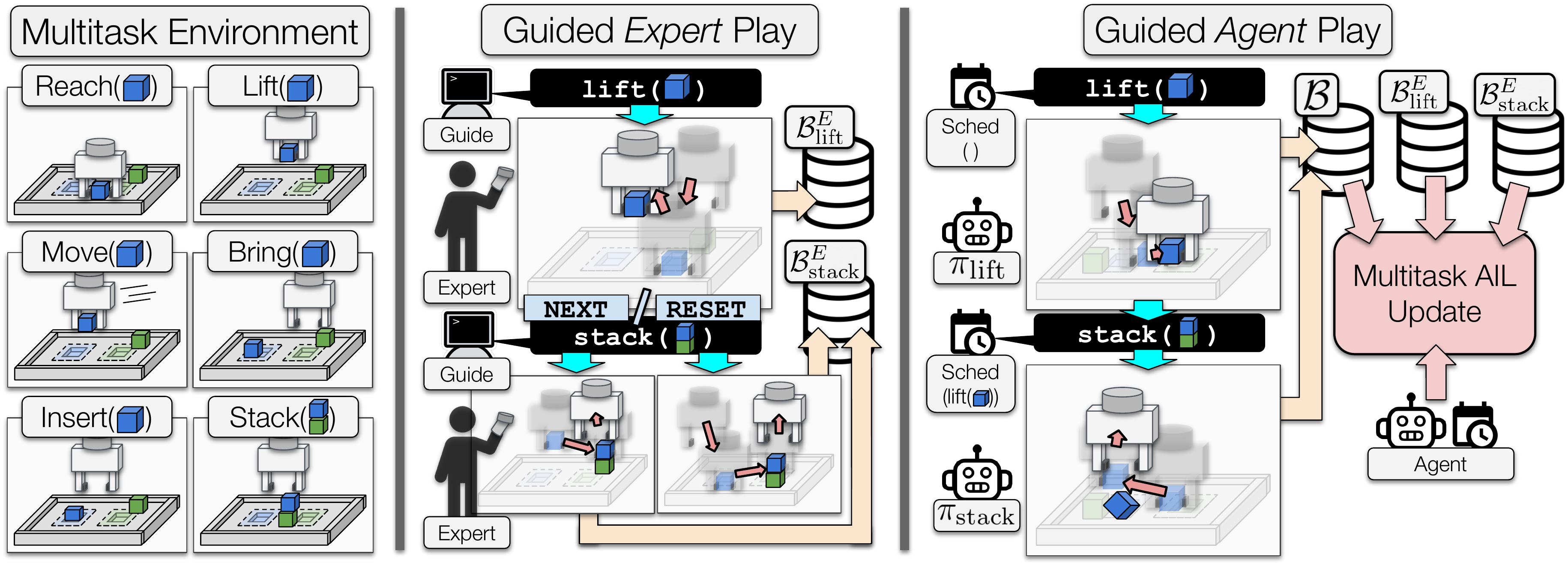}
	\caption{The main components of our system for learning from guided play. In a multitask environment, a guide prompts an expert for a mix of multitask demonstrations, after which we learn a multitask policy through scheduled hierarchical AIL.}
	\label{fig:system}
	\vspace{-4mm}
\end{figure*}

AIL has been shown to be effective in many continuous-control environments \cite{ho2016generative,kostrikov2019imitation}, but its application to long-horizon robotic manipulation tasks with a wide distribution of possible initial configurations remains challenging \cite{kostrikov2019imitation,orsini2021matters, rafailov2021visual}.
In this work, we investigate the use of AIL in a multitask robotic manipulation domain (see the left side of \Cref{fig:system}), specifically investigating stacking, unstacking and then stacking, reaching followed by bringing to a target, and reaching followed by inserting.
We find that a state-of-the-art AIL method, in which off-policy learning is used to maximize sample efficiency \cite{kostrikov2019imitation}, performs poorly on these complex manipulation tasks.
In all of our tasks, AIL even performs worse than the simpler alternative of supervised learning (often referred to as behavioural cloning (BC).
We hypothesize that this is because AIL, and more generally IRL, can provide something akin to a dense reward, but because this learned reward is not necessarily optimal for teaching, AIL alone may not necessarily enforce sufficiently diverse exploration to allow effective learning. 

To overcome this limitation of AIL, we present Learning from Guided Play (LfGP), in which we combine AIL with a scheduled approach to hierarchical RL (HRL), allowing an agent to ``play'' in the environment using the expert ``guide.''
Using expert demonstrations of multiple auxiliary tasks, along with a main task, we demonstrate that our scheduled hierarchical agent is able to learn much more effectively than with AIL alone.
This approach also affords several other benefits: learning a challenging main task becomes more efficient than supervised learning in terms of required expert data, no hand-crafted reward functions are needed as in \cite{riedmiller2018learning}, expert data on auxiliary tasks can be reused from one main task to another, and learning time can be reduced by reusing previously-learned auxiliary models on new main tasks through simple transfer learning.
We choose this name in reference to the work upon which we build \cite{riedmiller2018learning}, which is in turn inspired by the ``playful'' phase of learning experienced by children, as opposed to goal-directed learning.
In our case, \textit{guided} represents two separate but related ideas: first, the expert guides this play, as opposed to requiring hand-crafted sparse rewards as in \cite{riedmiller2018learning} (right side of \Cref{fig:system}), and second, the expert gathering of multitask, semi-structured demonstrations is \textit{guided} by uniform-random task selection (middle of \Cref{fig:system}), rather than requiring the expert to choose transitions between goals, as in \cite{lynch2020learning, guptaRelayPolicyLearning2019}.
Our specific contributions are:
\begin{enumerate}
    \item A novel approach to AIL that learns a reward and policy for a challenging main task by simultaneously learning rewards and policies for auxiliary tasks.
    \item A variety of manipulation experiments illustrating the viability of our approach, including comparisons with competing approaches to AIL and imitation learning.
    \item An extension of our method to the domain of transfer learning.
\end{enumerate}

\section{Related Work}

Imitation learning (IL) is often divided into two main categories: behavioural cloning (BC) \cite{pomerleau1989alvinn} and inverse reinforcement learning (IRL) \cite{ng2000algorithms,abbeel2004apprenticeship}. 
BC recovers the expert policy via supervised learning, but it suffers from compounding errors due to covariate shift \cite{ross2011reduction}.
Alternatively, IRL partially alleviates the covariate shift problem by estimating the reward function and then applying RL using the learned reward. 
A popular approach to IRL is adversarial imitation learning (AIL) \cite{ho2016generative,kostrikov2019imitation,DBLP:journals/corr/abs-1809-02925,hausman2018multi}, in which the expert policy is recovered by matching the occupancy measure between the generated data and the demonstration data. 
Our proposed method extends existing AIL algorithms by enabling more effective exploration via the use of a scheduled multitask model.

Agents learned via hierarchical reinforcement learning (HRL) methods, which aim to learn and act over multiple levels of temporal abstractions in long planning horizon tasks, are shown to provide more effective exploration than agents operating with only a single level \cite{riedmiller2018learning,sutton1999between,kulkarni2016hierarchical,nachum2018data,nachum2019does}. 
Our work most closely resembles hierarchical AIL methods that attempt to combine AIL with HRL \cite{hausman2018multi,henderson2018optiongan,sharma2018directed,jing2021adversarial}. 
Existing works \cite{henderson2018optiongan,sharma2018directed,jing2021adversarial} often formulate the hierarchical agent using the Options framework \cite{sutton1999between} and recover the reward function using AIL \cite{ho2016generative}. 
Both \cite{henderson2018optiongan} and \cite{jing2021adversarial} leverage task-specific expert demonstrations to learn options using mixture-of-experts and expectation-maximization strategies respectively. 
Our work focuses on expert demonstrations that include multiple tasks, each of which encodes clear semantic meaning. 
In the multitask setting, \cite{hausman2018multi} and \cite{sharma2018directed} leverage unsegmented, multitask expert demonstrations to learn low-level policies via a latent variable model.
Other work has used a large corpus of unsegmented but semantically meaningful ``play" expert data to bootstrap policy learning \cite{lynch2020learning, guptaRelayPolicyLearning2019}; we define our expert dataset as \textit{guided} play, in that the expert completes semantically meaningful auxiliary tasks with provided transitions, reducing the burden on the expert to generate this data arbitrarily and simultaneously providing auxiliary task labels. 
Compared with learning from unsegmented demonstrations, the use of segmented demonstrations, as in \cite{codevilla2018end}, ensures that we know which auxiliary tasks our model will be learning, and opens up the possibility of expert data reuse and transfer learning without further assumptions.
Finally, we deviate from the Options framework and build upon Scheduled Auxiliary Control (SAC-X) to train our hierarchical agent as it is shown to work well in complex manipulation tasks \cite{riedmiller2018learning,hertweck2020simple}.

\section{Problem Formulation}
We define a Markov Decision Process (MDP) as $\mdpm = \langle \mdps, \mdpa, R, \mdpp, \mdprho, \gamma \rangle$. 
The sets $\mdps$ and $\mdpa$ are respectively the state and action space, $R : S \times A \rightarrow \Real$ is a reward function, $\mdpp : S \times A \times S \rightarrow \Real_{\ge 0}$ is the state-transition environment dynamics distribution, $\mdprho : S \rightarrow \Real_{\ge 0}$ is the initial state distribution, and $\gamma$ is the discount factor. 
Actions are sampled from a stochastic policy $\pi(a|s)$.
The policy $\pi$ interacts with the environment to yield experience $\left( s_t, a_t, r_t, s_{t + 1} \right)$ for $t = 0, \dots, \infty$,  where $s_0 \sim \mdprho(\cdot), a_t \sim \pi(\cdot | s_t), s_{t+1} \sim \mdpp(\cdot | s_t, a_t), r_t = R(s_t, a_t)$.

We assume infinite-horizon, non-terminating environments where $t$ is unbounded for notational convenience, but our method also easily applies to the finite-horizon case.
We aim to learn a policy $\pi$ that maximizes the expected return
$J(\pi) = \mathbb{E}_{\pi}\left[ G(\tau_{0:\infty}) \right] = \mathbb{E}_\pi\left[ \sum_{t=0}^{\infty} \gamma^{t} R(s_t, a_t) \right]$,
where $\tau_{t:\infty} = \{(s_t, a_t), \dots\}$ is the trajectory starting in state $s_t$, and $G(\tau_{t:\infty})$ is the return of the trajectory. 

In this work, we focus on imitation learning (IL) in which $R$ is unknown and instead, we are given a finite set of expert demonstration $(s,a)$ pairs $\buffer^E = \left\{ (s,a)^{E,(1)}, \dots \right\}$.
In AIL, we attempt to simultaneously learn $\pi$ and a discriminator $D : S \times A \rightarrow \Real_{\ge 0}$ that differentiates between expert samples $(s,a)^E$ and policy samples $(s,a)^\pi$, and subsequently define $R$ using $D$ \cite{ho2016generative, kostrikov2019imitation}.
To accommodate hierarchical learning, we augment $\mdpm$ to contain auxiliary tasks, where $\tasks_{\text{aux}} = \left\{ \tasks_1, \dots, \tasks_K \right\}$ are separate MDPs that share $\mdps, \mdpa, \mdpp, \mdprho$ and $\gamma$ with the main task $\tasks_{\text{main}}$ but have their own reward functions $R_{k}$. 
With this modification, we refer to entities in our model that are specific to task $\tasks \in \tasks_{\text{all}}$, $\tasks_\text{all} = \tasks_\text{aux} \cup \left\{ \tasks_\text{main} \right\}$, as $(\cdot)_\tasks$.
We assume to have a set of expert data $\buffer^E_\tasks$ for each task.

\section{Learning from Guided Play (LfGP)}
\label{sec:method}
We now introduce Learning from Guided Play (LfGP), in which we improve the performance of off-policy AIL in complex manipulation tasks that require synthesizing a composition of potentially reusable skills. 
Hypothesizing that better exploration will improve performance \cite{nachum2018data}, our method specifically enforces the learning and exploration of these skills through a hierarchical framework \cite{riedmiller2018learning}, while still exclusively using expert data.
Our primary goal is to learn a policy $\pi_{\tasks_{\text{main}}}$ that can solve the main task $\tasks_{\text{main}}$, with a secondary goal of also learning auxiliary task policies $\pi_{\tasks_{\text{1}}}, \dots, \pi_{\tasks_{\text{K}}}$ that can be used both for improved exploration and for transfer learning.
More specifically, we derive a hierarchical learning objective that is decomposed into three parts: i) recovering the reward function of each task with expert demonstrations, ii) training all policies to achieve their respective goals, and iii) using all policies for effective exploration in $\tasks_\text{main}$. 
For a summary of the algorithm, see our supplementary material.

\subsection{Learning the Reward Function}
We first describe how to recover the reward functions from expert demonstrations. 
For each task $\tasks \in \tasks_\text{all}$, we learn a discriminator $D_{\tasks}(s, a)$ that is used to define the reward function for policy optimization. 
The discriminator is trained in an off-policy manner \cite{DBLP:journals/corr/abs-1809-02925}:
\begin{eqnarray}
\begin{split}
    \max_{D_\tasks} \mathbb{E}_{\buffer}\left[ \frac{\rho_{\pi_\tasks}(s, a)}{\rho_{\buffer}(s, a)} \log \left( 1 - D_\tasks(s, a)\right) \right] + \mathbb{E}_{\buffer^E_{\tasks}}\left[ \log \left(D_\tasks(s, a)\right) \right],
\end{split}
\label{eqn:GAIL}
\end{eqnarray}
where $\buffer$ is the replay buffer storing transitions collected while interacting with the environment, $\rho_{\pi_\tasks}(s, a)$ the intention occupancy measure, and $\rho_{\buffer}$ the distribution induced by the replay buffer $\buffer$. 
In practice, following \cite{DBLP:journals/corr/abs-1809-02925}, we sample the transitions from $\buffer$ without applying importance sampling:
\begin{eqnarray}
    \label{DAC_og}
    \max_{D_\tasks} \mathbb{E}_{\buffer}\left[ \log \left( 1 - D_\tasks(s, a)\right) \right] + \mathbb{E}_{\buffer^E_{\tasks}}\left[ \log \left( D_\tasks(s, a)\right) \right].
\end{eqnarray}

Each discriminator tries to differentiate the occupancy measure between the distributions induced by $\buffer^E_\tasks$ and $\buffer$. 
As a result, the policy $\pi_\tasks$ attempts to confuse the discriminator by minimizing Equation \ref{DAC_og} through changing the distribution induced by the replay buffer to be similar to the expert demonstrations. 
In our implementation, the discriminators minimize the joint discriminator loss
\begin{equation}
\label{DAC}
\begin{aligned}
    \mathcal{L}(D) = -\sum_{\tasks}^{\tasks_\text{all}} \mathbb{E}_{\buffer}\left[ \log \left( 1 - D_\tasks(s, a)\right) \right] + \mathbb{E}_{\buffer^E_{\tasks}}\left[ \log \left( D_\tasks(s, a)\right) \right].
\end{aligned}
\end{equation}
The resulting $D_\tasks$ can be used to define various reward functions \cite{DBLP:journals/corr/abs-1809-02925}---we define the reward function for each task $\tasks$ to be $D_\tasks$. Thus, we define the expected return for $\pi_\tasks$ as
\begin{equation}
    J(\pi_\tasks) = 
    \mathbb{E}_{\pi_\tasks}\left[ G_\tasks(\tau_{t:\infty}) \right] = 
    \mathbb{E}_{\pi_\tasks}\left[ \sum_{t = 0}^{\infty} \gamma^{t} D_\tasks(s_t, a_t) \right].
\label{task_return}
\end{equation}

\subsection{Learning the Hierarchical Agent}
We adapt SAC-X \cite{riedmiller2018learning} to learn the hierarchical agent. 
The agent includes two levels of policies: intention policies (low-level) and a scheduler (high-level), as well as the Q-functions and the discriminators.
The intentions aim to solve their corresponding tasks (i.e. intentions are $\pi_\tasks$ aiming to maximize task return $J(\pi_\tasks)$), whereas the scheduler aims to maximize the expected return for $\tasks_\text{main}$ by selecting a sequence of intentions to interact with the environment.
For the remainder of this paper, when we refer to a policy, we are referring to an intention policy, as opposed to the scheduler, unless otherwise specified.

\subsubsection{Learning the Intentions}
We learn each intention using Soft Actor-Critic (SAC) \cite{haarnoja2018soft}, an actor-critic algorithm that optimizes for the entropy-regularized objective, though any off-policy RL algorithm would suffice.
We augment Equation \ref{task_return} with entropy regularization as
\begin{eqnarray}
    J(\pi_\tasks) = 
    \mathbb{E}_{\pi_\tasks}\left[ \sum_{t = 0}^{\infty} \gamma^{t} (D_\tasks(s_t, a_t) + \alpha\mathcal{H}(\pi_\tasks(\cdot | s_t))) \right],
\end{eqnarray}
where the learned temperature $\alpha$ determines the importance of the entropy term against the reward, and $\mathcal{H}(\pi_\tasks (\cdot | s_t))$ is the entropy of the intention $\pi_\tasks$ at state $s_t$. 
The corresponding soft Q-function $Q_\tasks(s_t, a_t)$ for task $\tasks$ is defined as 
\begin{equation}
\begin{aligned}
    Q_\tasks&(s_t, a_t) = D_\tasks(s_t, a_t)  + \mathbb{E}_{\pi_\tasks}\left[ \sum_{t = 0}^{\infty} \gamma^t (D_\tasks(s_{t+1}, a_{t+1}) + \alpha\mathcal{H}(\pi_\tasks(\cdot | s_{t+1}))) \right].
\end{aligned}
\label{q_func}
\end{equation}
The intentions maximize the joint policy improvement objective:
\begin{equation}
    \label{complete_PI}
    \mathcal{L}(\pi) = \sum_{\tasks}^{\tasks_\text{all}}\mathbb{E}_{s \sim \buffer, a \sim \pi_\tasks(\cdot | s)}\left[ Q_\tasks(s, a) - \alpha \log \pi_\tasks(a | s) \right].
\end{equation}
Notice that the replay buffer $\buffer$ contains transitions generated using different intentions. This allows each intention $\pi_\tasks$ to learn optimal actions for its respective task starting at any state visited by any intention. 
In other words, other intentions $\pi_{\tasks' \ne \tasks}$ interacting with the environment provide exploration samples for $\pi_\tasks$.

For policy evaluation, the soft Q-functions $Q_\tasks$ for each $\pi_\tasks$ minimize the joint soft Bellman residual
\begin{equation}
    \label{complete_PE}
    \mathcal{L}(Q) = \sum_{\tasks}^{\tasks_\text{all}}\mathbb{E}_{(s, a, r, s') \sim \buffer, a' \sim \pi_\tasks(\cdot | s')}\left[ (Q_\tasks(s, a) - \delta_\tasks)^2 \right],
\end{equation}
\begin{equation}
    \delta_\tasks = r + \gamma Q_\tasks(s', a') - \alpha \log \pi_\tasks(a' | s').
\end{equation}

\subsubsection{Learning the Scheduler}
The scheduler composes a sequence of intentions to interact with the environment. 
The scheduler can switch between intentions every $\xi$ timesteps. 
With $\gamma < 1$, we define $H$ as the total number of possible intention switches within an episode.\footnote{We can choose $H$ in the infinite horizon setting using the effective horizon $\frac{1}{1 - \gamma}$.} 
The $H$ intention choices made within the episode are defined as $\tasks^{0:H-1} = \left\{ \tasks^{(0)}, \dots, \tasks^{(H-1)} \right\}$, where $\tasks^{(h)} \in \tasks_\text{all}$. 
The main task's return given the intention choices is then defined as
\begin{eqnarray}
    G_{\tasks_\text{main}}(\tasks^{0:H-1}) = \sum_{h=0}^{H - 1} \sum_{t=h\xi}^{(h + 1)\xi - 1} \gamma^t D_{\tasks_\text{main}}(s_t, a_t),
\end{eqnarray}
where $a_t \sim \pi_{\tasks^{(h)}}(\cdot | s_t)$ is the action taken at timestep $t$, sampled from the chosen intention $\tasks^{(h)}$ in the $h^\text{th}$ scheduler period. 
We define the Q-function for the scheduler as $Q_S(\tasks^{0:h-1}, \tasks^{(h)}) = \mathbb{E}_{P_S}\left[ G_{\tasks_\text{main}}(\tasks^{h:H-1}) | \tasks^{0:h-1} \right]$ and represent the scheduler as a Boltzmann distribution $P_S$ over the Q-function. The scheduler maximizes the expected return of the main task following the scheduler:
\begin{eqnarray}
    \mathcal{L}(S) = \mathbb{E}_{P_S}\left[ Q_S(\emptyset, \tasks^{(0)}) \right].
    \label{scheduler}
\end{eqnarray}
We use Monte Carlo returns to estimate $Q_S$, estimating the expected return using the exponential moving average
\begin{equation}
\begin{aligned}
    \label{EMA}
    Q_S(\tasks^{0:h-1}, \tasks^{(h)}) = (1 - \phi)Q_S(\tasks^{0:h-1}, \tasks^{(h)}) + \phi G_{\tasks_\text{main}}(\tasks^{h:H}),
\end{aligned}
\end{equation}
where $\phi \in [0, 1]$ represents the amount of discounting on older returns and $G_{\tasks_\text{main}}(\tasks^{h:H})$ is the cumulative discounted return of the trajectory $\tau$ starting at timestep $h\xi$.

\subsection{Expert Data Collection} \label{sec:exp_data_collection}
We assume that each $\tasks \in \tasks_\text{all}$ has, for evaluation purposes only, a binary indicator of success.
In single-task imitation learning (IL) where this assumption is valid, expert data is typically collected by allowing the expert to control the agent until success conditions are met.
At that point, the environment is reset following $\rho_0$ and collection is repeated for a fixed number of episodes or $(s,a)$ pairs.
In LfGP, and more generally for multitask IL, expert data can also be collected this way for each $\tasks$ separately: we refer to this strategy, which we use for most of our experiments, as \textit{reset-based} expert data collection.

A limitation of reset-based expert data collection is that, when training LfGP, the initial state of each individual policy can be any state that $\mathcal{M}$ has been left in by a previous policy, which may include states not in the distribution of $\rho_0$.
This ``transition'' initial state distribution, which we call $\rho_0(s | \tasks')$, where $\tasks' \in \tasks_{\text{all}}$ corresponds to the previously running policy $\pi_{\tasks'}$, would be challenging to sample from---it relies on the policies, and it may include states which are impractical to manually reset to (e.g. objects may start off as grasped or in mid-air).
Consequently, we propose an alternative data collection strategy: \textit{play-based} expert data collection, where we alternate between uniformly sampling the next task for an expert to complete and having the expert execute that task until success. 
In our implementation, we also reset the environment following $\rho_0$ periodically.
See middle section of \Cref{fig:system} for a comparison of the two methods.

\section{Experiments} \label{sec:exp}
In our experiments, we attempt to answer the following questions:
\begin{enumerate}
    \item How does the performance of LfGP compare with other existing state-of-the-art IL methods in challenging manipulation tasks, in terms of success rate, environment sample efficiency, and expert sample efficiency?
    \item Are models trained using LfGP able to transfer from one main task to another? Does this transfer afford any benefits over learning the new main task from scratch?
    \item Does the use of play-based expert data improve performance (see \Cref{sec:exp_data_collection})?
\end{enumerate}

\subsection{Experimental Setup} \label{sec:exp_setup}
\begin{figure*}[t]
    \vspace{-3mm}
	\centering
	\begin{subfigure}{0.49\textwidth}
	    \includegraphics[width=\textwidth]{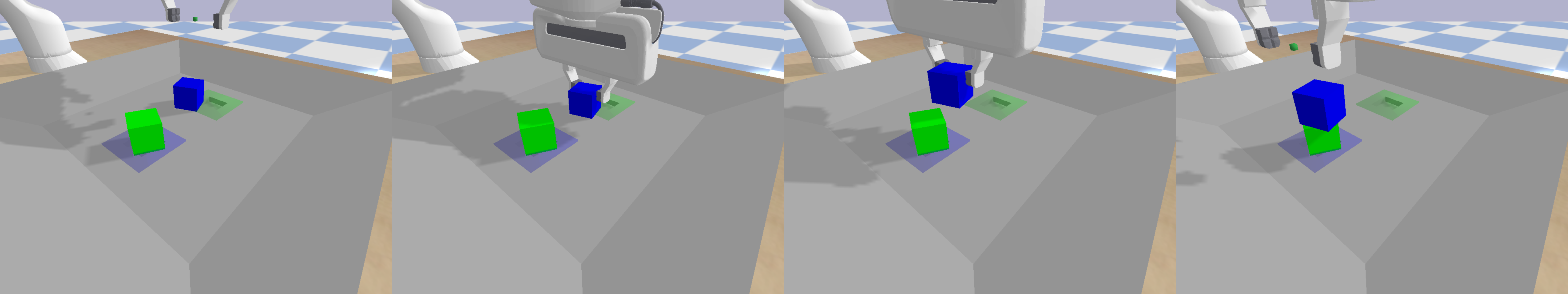}
	\end{subfigure}
	\hspace{1mm}
	\begin{subfigure}{0.49\textwidth}
	    \includegraphics[width=\textwidth]{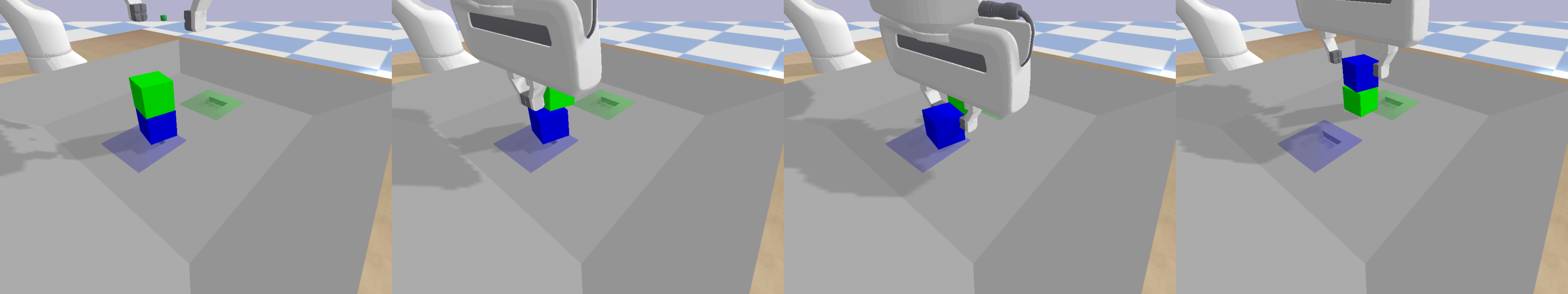}
	\end{subfigure}
	\par \vspace{2mm}
	\begin{subfigure}{0.49\textwidth}
	    \includegraphics[width=\textwidth]{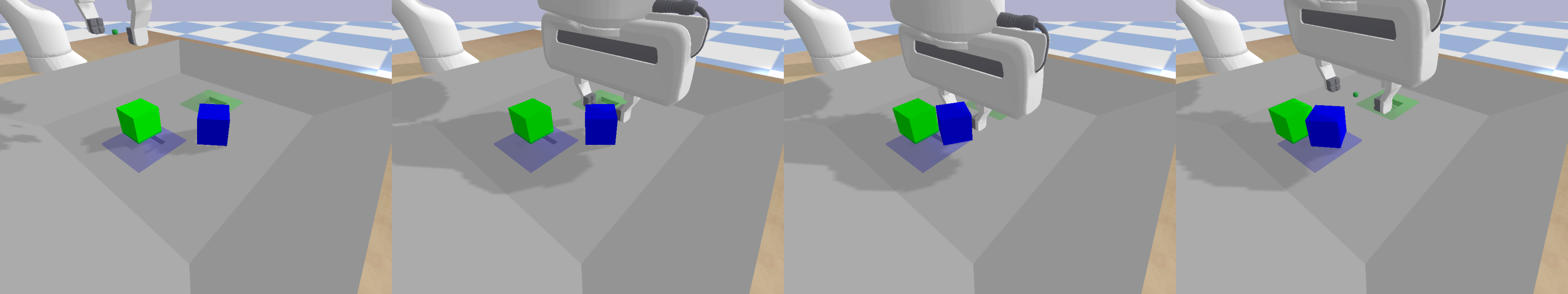}
	\end{subfigure}
	\hspace{1mm}
	\begin{subfigure}{0.49\textwidth}
	    \includegraphics[width=\textwidth]{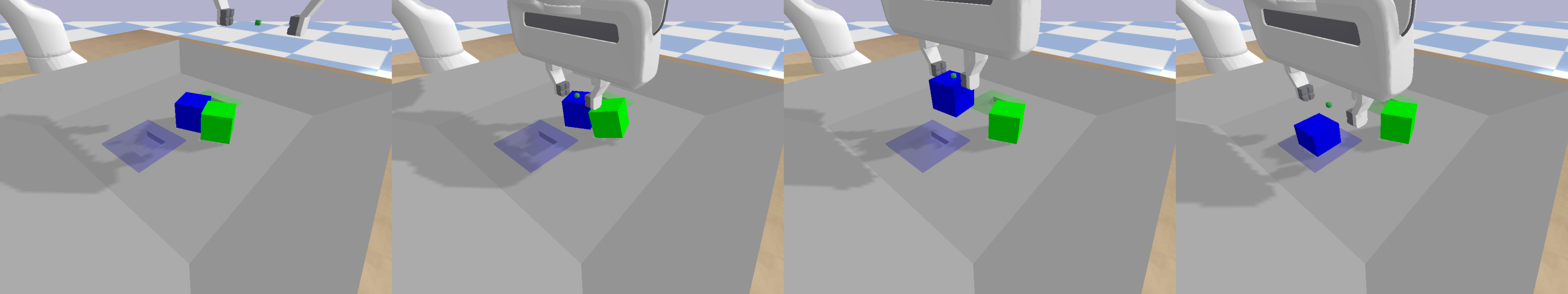}
	\end{subfigure}
	
	\caption{Example successful runs of our four main tasks. Left to right, top to bottom: Stack, Unstack-Stack, Bring, Insert.}
	\label{fig:example_runs}
\end{figure*}

We begin by describing our experimental setup.
We complete experiments in a simulation environment containing a 7-DOF Franka Emika Panda manipulator, one green and one blue block, fixed zones corresponding to the green and blue blocks, and one slot in each zone with $<1$mm tolerance for fitting the blocks (see bottom right of Figure \ref{fig:example_runs}).
The environment is designed such that several different challenging tasks can be completed with a common observation and action space.
The main tasks that we investigate are Stack, Unstack-Stack, Bring, and Insert (see \Cref{fig:example_runs} for examples).
When learning Unstack-Stack, the green block is initialized on top of the blue block.
For more details on our environment and definitions of task success, see \Cref{sec:env_details} and \Cref{sec:evaluation} respectively.
We also define a set of auxiliary tasks: Open-Gripper, Close-Gripper, Reach, Lift, Move-Object, and Bring (Bring is both a main task and an auxiliary task for Insert), all of which have some degree of re-usability between main tasks.
Table \ref{tab:env_details} shows the relationship between main and auxiliary tasks.

We implemented several multitask and single-task baselines to compare our method to.
The multitask algorithms, like our method, simultaneously attempt to learn to complete the main task as well as the auxiliary tasks.
The single-task algorithms only attempt to learn to complete the main task.
In general, we consider that a multitask algorithm is more useful than a single-task algorithm, given the potential to reuse trained models for learning new tasks, as we show in Section \ref{sec:transfer_exp}.

The first multitask algorithm that we compare to is a ``No Schedule'' version of our method (LfGP-NS), where the models are trained similarly to LfGP, but the scheduler only chooses the main task; this is meant to be a similar baseline to the Intentional Unintentional Agent shown in \cite{riedmiller2018learning, cabi2017intentional}.
The expected benefit is that the policies, the Q-functions, and the discriminators share their initial layers between tasks, so the auxiliary tasks can accelerate learning the representation for the main-task models \cite{jaderberg2016reinforcement}.
We also compare to a multitask variant of behavioural cloning (BC), in which we simultaneously train every task by minimizing a multitask mean squared error objective
\begin{align}
    \mathcal{L}(\pi) = \sum_{\tasks}^{\tasks_\text{all}} \sum_{(s,a) \in \buffer^E_{\tasks}} \left(\pi_{\tasks}(s) - a \right)^2,
\end{align}
taking only the mean action outputs from each $\pi_{\tasks}$.

The single-task algorithms we compare to are BC and Discriminator-Actor-Critic (DAC) \cite{kostrikov2019imitation}.
To make a fairer comparison and evaluate our method's expert sample efficiency, we provide BC and DAC with an equivalent amount of \textit{total} expert data as our multitask methods as shown in Table \ref{tab:env_details}.
To show the possible benefits or detriments of learning only a single main task with an equivalent amount of main task data as the multitask algorithm, we also show results for single-task BC with an equivalent amount of main task data to the multitask algorithms, which we call BC (less data).

We gathered expert data by first training an expert policy using SAC-X \cite{riedmiller2018learning} (see \cref{sec:expert_gen}).
We then ran the expert policies to collect various amounts of expert data as described in Table \ref{tab:env_details}.
In these experiments, in the interest of testing the benefits of having reusable expert data, we use reset-based expert data (see \Cref{sec:exp_data_collection}).

\begin{table}
	\centering
	\small
	\caption{A summary of the reusability of the expert data used to generate the performance results described in Section \ref{sec:perf_res}. Each letter under ``Dataset Sizes'' is the first letter of a single (auxiliary) task, and bolded letters indicate that a dataset was reused for more than one main task (e.g., \textbf{O}pen-Gripper was used for all four main tasks). The numbers in this table refer to $(s,a)$ pairs.}
	\begin{tabularx}{.9\textwidth}{lllXXX}
	    \toprule
		Algorithm Class                                           &     Task & Dataset Sizes & Reusable & Single & Total
		\\\midrule
		\textit{Multitask}                                            & Stack & \textbf{OC}S\textbf{RLM}: 9k/task          & \textbf{45k}              & 9k                  & 54k   \\
                                                             & U-Stack & \textbf{OC}U\textbf{RLM}: 9k/task          & \textbf{45k}              & 9k                  & 54k   \\
		                                                           & Bring & \textbf{OCBRLM}: 9k/task          & \textbf{54k}              & 0                   & 54k   \\
		                                                           & Insert    & \textbf{OC}I\textbf{BRLM}: 9k/task         & \textbf{54k}              & 9k                  & 63k   \\ \midrule
		\textit{Single-task}                                            & Stack      & S: 54k (less: 9k)       & 0                & 54k (9k)                  & 54k (9k)  \\
		                                                       & U-Stack     & U: 54k (less: 9k)        & 0                & 54k (9k)                & 54k (9k)  \\
		                                                           & Bring     & B: 54k (less: 9k)        & 0                & 54k (9k)                & 54k (9k)  \\
		                                                           & Insert    & I: 63k (less: 9k)        & 0                & 63k (9k)                & 63k (9k)  \\
	\end{tabularx}
	\label{tab:env_details}
\end{table}

\begin{figure*}[t]
	\centering
	\includegraphics[width=.95\textwidth]{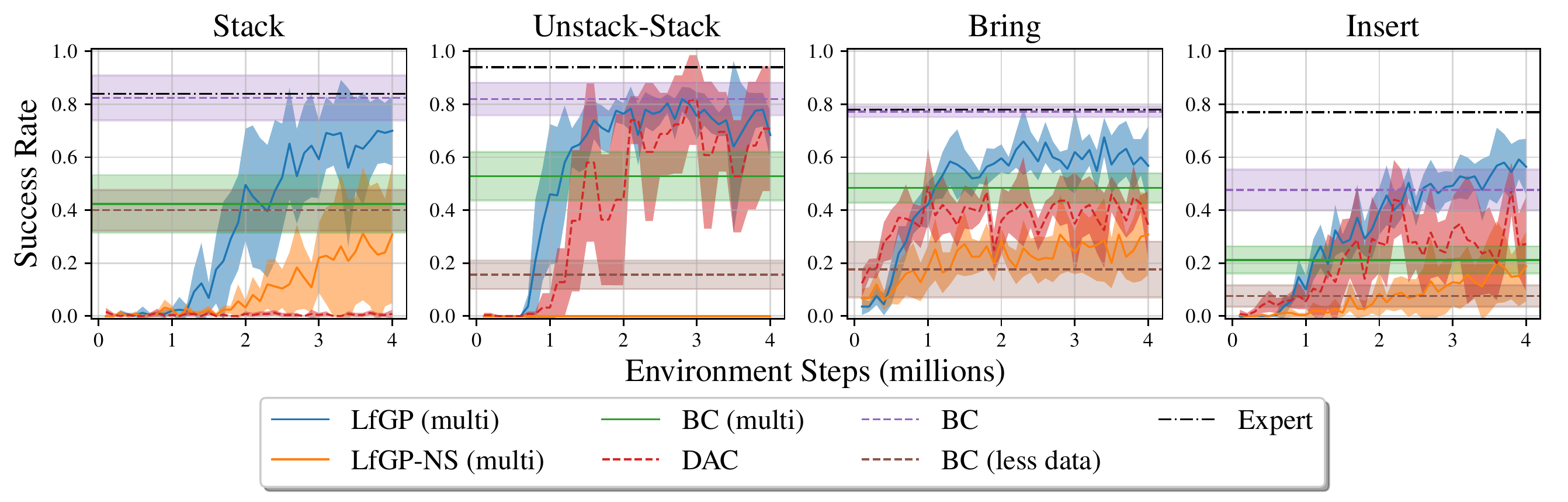}
	\caption{Performance results for our method (LfGP) and the baselines, as described in Section \ref{sec:exp_setup}. Solid lines are multitask methods, dotted lines are single-task methods, and shaded area corresponds to standard deviation across five seeds.}
	\label{fig:perf_results}
	\vspace{-3mm}
\end{figure*}
\begin{figure*}[t]
	\centering
	\includegraphics[width=\textwidth]{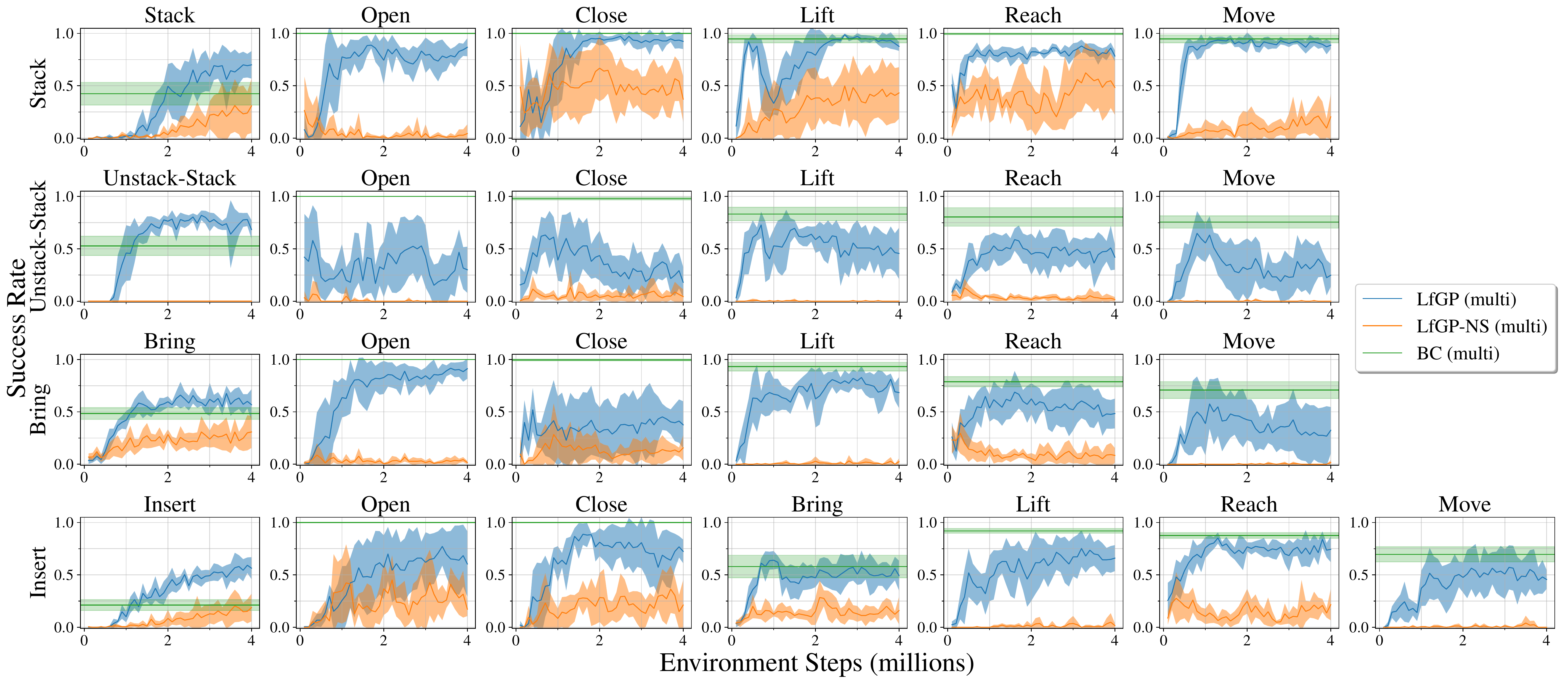}
	\caption{Performance for LfGP and the multitask baselines across all tasks, shaded area corresponds to standard deviation.}
	\label{fig:multi_perf_results}
	\vspace{-5mm}
\end{figure*}

\subsection{Performance Results} \label{sec:perf_res}
For the AIL algorithms, we freeze the policies every 100k interactions and evaluate those policies for 50 randomized episodes, only using the mean action outputs. 
We trained each agent to 4 million environment steps.
For the BC algorithms, we train the policies by dividing each expert buffer into a 70\%/30\% train/validation split, taking the policy after validation error has not improved for 100 epochs and evaluating it on 50 randomized episodes.
For all algorithms, we test across five seeds and report the mean and standard deviation of all seeds.
For more details of our training algorithms, including hyperparameters, see our supplementary material.
The performance results for all methods with respect to the main tasks we tested are shown in Figure \ref{fig:perf_results}.

It is clear that our method outperforms multitask BC and LfGP-NS, demonstrating the necessity of exploring each of the policies to maximize performance.
Intriguingly, even with substantially more main-task expert data, DAC alone is unable to match the performance of our method, and completely fails in the Stacking task.
Single-task BC with more main-task data outperforms our method in Stack and Bring, but we emphasize that this baseline has two major limitations compared with ours: none of the collected expert data for this policy is reusable for other tasks, and the policy itself does not have any learned auxiliary policies that could be reused.

The performance results for all multitask methods and all auxiliary tasks are shown in Figure \ref{fig:multi_perf_results}.
One notable finding is that multitask BC tends to perform quite well on the auxiliary tasks, and, in fact, for all auxiliary tasks, outperforms LfGP. 
We suspect this is because, compared to the main task, the auxiliary tasks are shorter horizon and simpler than the main task, making them very good candidates for BC.
Furthermore, since LfGP maximizes the expected return of the main task, LfGP does not necessarily perform as well on the auxiliary tasks compared to multitask BC, which simultaneously matches the expert for all tasks.
A natural followup question is whether we could combine the benefits of quick learning using multitask BC on the simpler auxiliary tasks with the improved performance of LfGP on main tasks---our preliminary results were negative so we leave this as future work.

\subsection{Transfer Learning Results} \label{sec:transfer_exp}
\begin{figure}[t]
	\centering
	\begin{subfigure}{0.45\textwidth}
	    \includegraphics[width=\textwidth]{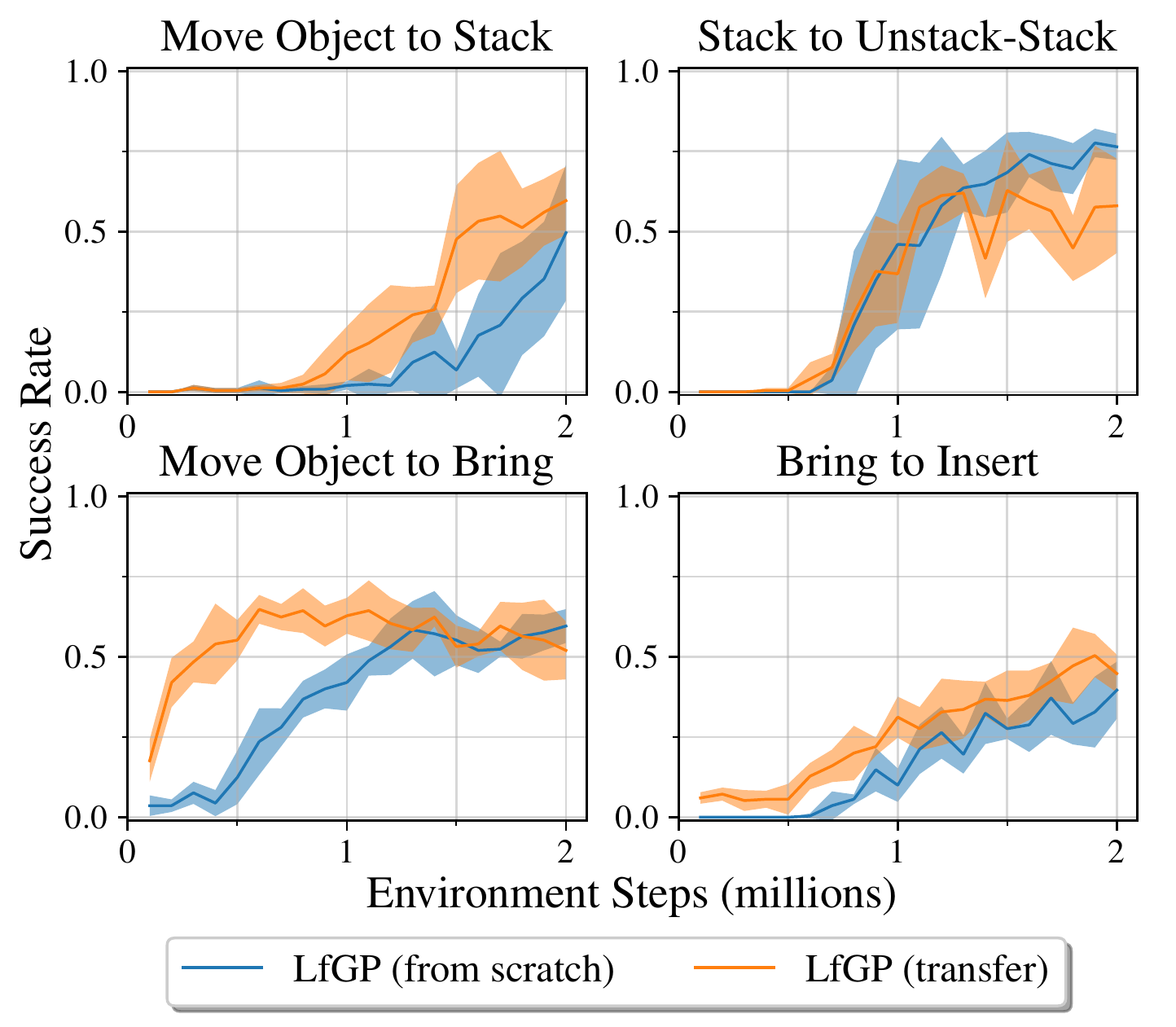}
	\end{subfigure}
	\hfill
	\begin{subfigure}{0.45\textwidth}
	    \includegraphics[width=\textwidth]{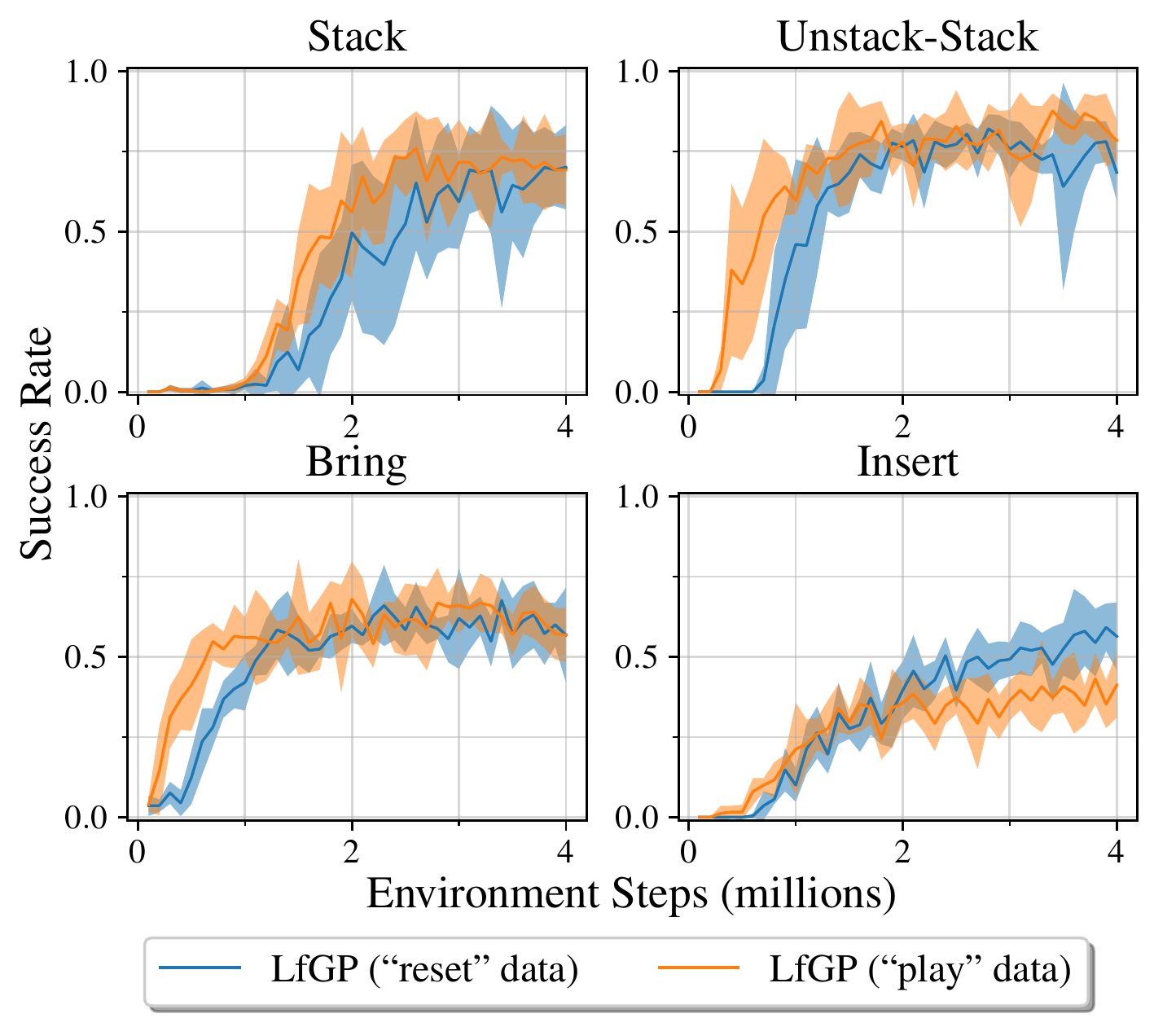}
	\end{subfigure}
	\caption{\textbf{Left:} The results of our transfer experiments. \textbf{Right:} The results of our play-based expert data experiments. Shaded area corresponds to standard deviation. }
	\label{fig:transfer_and_play_results}
\end{figure}
In this section, we show the results of experiments in which we attempt to reuse policies that learned to complete one main task to complete another main task.
Specifically, we are interested in reducing the learning time for a main task compared with learning from scratch.
To complete these experiments, we chose a relatively simple formulation as a proof-of-concept.
For each main task, we take a high-performing existing model that uses a subset of auxiliary tasks required for the new main task and save the parameters and the replay buffer of the existing model. 
We then train a new model on the new task by loading the saved parameters and replay buffer, adding new model outputs for the new main task.
For Stack and Bring, we transfer from a model with $\tasks_{\text{main}} = \text{Move-Object}$ and $\tasks_{\text{aux}} = \left\{ \text{Open-Gripper, Close-Gripper, Reach, Lift} \right\}$.
For Unstack-Stack and Insert, we use $\tasks_{\text{main}} = \text{Stack}$ and $\tasks_{\text{main}} = \text{Bring}$ models respectively, with the same $\tasks_{\text{aux}}$ as in \Cref{tab:env_details}.

The results of these experiments are shown on the left side of Figure \ref{fig:transfer_and_play_results}.
It is clear that in all cases except for Unstack-Stack, transferring trained models from one main task to another is more sample efficient than from scratch. 
This result is particularly clear in the case of Bring.
For Unstack-Stack, training speed remains roughly consistent, but final performance is actually considerably lower with the transfer model.
Given that this experiment was only meant to show that transfer learning would be possible and could provide benefits in some cases, we leave investigating this issue for future work.

\subsection{Expert Data Collection Schemes}
As described in \Cref{sec:exp_data_collection}, there are (at least) two strategies for collecting expert data to be used with LfGP.
In this section, we compare the training performance of LfGP between reset-based and play-based expert data.

We show the comparison between the results of training LfGP with reset-based and with play-based data on the right side of Figure \ref{fig:transfer_and_play_results}. 
For Stack, Unstack-Stack, and Bring, play-based data appears to generally increase the learning speed of LfGP, implying that matching the transition distribution appears to be beneficial for learning, although there is no significant effect on final performance.
Conversely, for Insert, play-based data appears to have only a marginal effect on learning speed, while having fairly significant negative impact on final performance.
This could be because the Insert task is the least forgiving in terms of the required final state of the object, and reset-based data may contain more transitions between near-insertions and complete insertions than play-based data.

Compared with reset-based data, while play-based data assures that the expert distribution better matches the learning distribution, it also has the downside of making it harder to reuse.
In the case of reset-based data, one can easily add a new dataset corresponding to a new task, while keeping existing datasets the same.
In play-based, and in our experiments, each individual main task has its own dataset, given that the ``transition" initial state distribution should contain data from $\tasks_{\text{all}}$, which changes depending on $\tasks_{\text{main}}$.

\section{Conclusion}
In this work, we identified and addressed an exploration problem in adversarial imitation learning (AIL) algorithms applied to manipulation tasks by introducing Learning from Guided Play (LfGP), a scheduled hierarchical AIL method that enables more effective exploration through the use of ``playful'' scheduled auxiliary tasks, ``guided'' by expert data, as opposed to hand-crafted rewards. 
We demonstrated that our method substantially outperforms an existing state-of-the-art AIL algorithm.
As well, with significantly less main task data, our algorithm can perform comparably to behaviour cloning (BC) and significantly outperforms multitask BC given the same data.
Furthermore, our method can leverage reusable expert data and its learned intentions can be applied to transfer learning.
As future work, we intend to further investigate alternative approaches to transfer learning, such as a combination of multitask BC with our method, to see if the complementary advantages of supervised learning and AIL can reduce the number of required environment interactions.

\section*{Broader Impact}
In this paper, we introduced a scheduled hierarchical approach to adversarial imitation learning (AIL), and demonstrated its successful application to a simulated robotic manipulation domain.
Robotic manipulation systems are inherently dangerous, though significant strides have been made in the field of \textit{collaborative} robotics in the past few decades \cite{colgateCobotsRobotsCollaboration1996}, in which lighter arms with speed and force limitations are used in close proximity with humans.
The simulated arm from this work, the Franka Emika Panda, is an example of such an arm.
Like any other reinforcement learning (RL) method applied to a physical system, ours shares the safety concern that a system that exhibits any degree of random exploration can act in unexpected ways, potentially causing harm through damage to the agent itself, the environment, or to present humans.
In our method, this could be exacerbated by the abrupt switching from one task to another; when using this approach on a real platform, care should be taken to constrain the robot's movements through soft and/or hard boundaries.
This abruptness could also be reduced by using some kind of soft switching mechanism between hierarchical members \cite{kipf2019compositional, lee2019composing}.
Compared with RL, our method could also learn unexpected behaviours due to the existence of suboptimal expert data, but this is a well-studied issue in the field of IL \cite{aroraSurveyInverseReinforcement2018}, so there is no shortage of approaches that could be combined with ours to address potentially imperfect demonstrations.

\bibliographystyle{unsrt}
\bibliography{references}

\begin{thebibliography}{10}

\bibitem{sutton2018reinforcement}
Richard~S Sutton and Andrew~G Barto.
\newblock {\em Reinforcement learning: An introduction}.
\newblock MIT press, 2018.

\bibitem{bellemareUnifyingCountBasedExploration2016b}
Marc Bellemare, Sriram Srinivasan, Georg Ostrovski, Tom Schaul, David Saxton,
  and Remi Munos.
\newblock Unifying {{Count}}-{{Based Exploration}} and {{Intrinsic
  Motivation}}.
\newblock In {\em Advances in {{Neural Information Processing Systems}}},
  volume~29. {Curran Associates, Inc.}, 2016.

\bibitem{nairOvercomingExplorationReinforcement2018}
Ashvin Nair, Bob McGrew, Marcin Andrychowicz, Wojciech Zaremba, and Pieter
  Abbeel.
\newblock Overcoming {{Exploration}} in {{Reinforcement Learning}} with
  {{Demonstrations}}.
\newblock In {\em 2018 {{IEEE International Conference}} on {{Robotics}} and
  {{Automation}} ({{ICRA}})}, pages 6292--6299, 2018.

\bibitem{ngShapingPolicySearch2003a}
Andrew~Y. Ng and Michael~I. Jordan.
\newblock {\em Shaping and Policy Search in Reinforcement Learning}.
\newblock PhD thesis, University of California, Berkeley, 2003.

\bibitem{ho2016generative}
Jonathan Ho and Stefano Ermon.
\newblock Generative adversarial imitation learning.
\newblock {\em Advances in neural information processing systems},
  29:4565--4573, 2016.

\bibitem{fu2018learning}
Justin Fu, Katie Luo, and Sergey Levine.
\newblock Learning robust rewards with adverserial inverse reinforcement
  learning.
\newblock In {\em International Conference on Learning Representations}, 2018.

\bibitem{kostrikov2019imitation}
Ilya Kostrikov, Ofir Nachum, and Jonathan Tompson.
\newblock Imitation learning via off-policy distribution matching.
\newblock In {\em International Conference on Learning Representations}, 2019.

\bibitem{orsini2021matters}
Manu Orsini, Anton Raichuk, L{\'e}onard Hussenot, Damien Vincent, Robert
  Dadashi, Sertan Girgin, Matthieu Geist, Olivier Bachem, Olivier Pietquin, and
  Marcin Andrychowicz.
\newblock What matters for adversarial imitation learning?
\newblock {\em arXiv preprint arXiv:2106.00672}, 2021.

\bibitem{rafailov2021visual}
Rafael Rafailov, Tianhe Yu, Aravind Rajeswaran, and Chelsea Finn.
\newblock Visual adversarial imitation learning using variational models.
\newblock In {\em ICML 2021 Workshop on Unsupervised Reinforcement Learning},
  2021.

\bibitem{riedmiller2018learning}
Martin Riedmiller, Roland Hafner, Thomas Lampe, Michael Neunert, Jonas Degrave,
  Tom Wiele, Vlad Mnih, Nicolas Heess, and Jost~Tobias Springenberg.
\newblock Learning by playing solving sparse reward tasks from scratch.
\newblock In {\em International Conference on Machine Learning}, pages
  4344--4353. PMLR, 2018.

\bibitem{lynch2020learning}
Corey Lynch, Mohi Khansari, Ted Xiao, Vikash Kumar, Jonathan Tompson, Sergey
  Levine, and Pierre Sermanet.
\newblock Learning latent plans from play.
\newblock In {\em Conference on Robot Learning}, pages 1113--1132. PMLR, 2020.

\bibitem{guptaRelayPolicyLearning2019}
Abhishek Gupta, Vikash Kumar, Corey Lynch, Sergey Levine, and Karol Hausman.
\newblock Relay policy learning: Solving long horizon tasks via imitation and
  reinforcement learning.
\newblock In {\em Conference on {{Robot Learning}} ({{CoRL}})}, 2019.

\bibitem{pomerleau1989alvinn}
Dean~A. Pomerleau.
\newblock {{ALVINN}}: {{An Autonomous Land Vehicle}} in a {{Neural Network}}.
\newblock In D.~S. Touretzky, editor, {\em Proceedings of the Annual Conference
  on Neural Information Processing Systems ({{NIPS}}'89)}, pages 305--313,
  {Denver, CO, USA}, 1989. {Morgan-Kaufmann}.

\bibitem{ng2000algorithms}
Andrew Ng and Stuart Russell.
\newblock Algorithms for inverse reinforcement learning.
\newblock {\em Proceedings of the Seventeenth International Conference on
  Machine Learning}, 0:663--670, 2000.

\bibitem{abbeel2004apprenticeship}
Pieter Abbeel and Andrew~Y Ng.
\newblock Apprenticeship learning via inverse reinforcement learning.
\newblock In {\em Proceedings of the twenty-first international conference on
  Machine learning}, page~1, 2004.

\bibitem{ross2011reduction}
St{\'e}phane Ross, Geoffrey Gordon, and Drew Bagnell.
\newblock A reduction of imitation learning and structured prediction to
  no-regret online learning.
\newblock In {\em Proceedings of the fourteenth international conference on
  artificial intelligence and statistics}, pages 627--635. JMLR Workshop and
  Conference Proceedings, 2011.

\bibitem{DBLP:journals/corr/abs-1809-02925}
Ilya Kostrikov, Kumar~Krishna Agrawal, Sergey Levine, and Jonathan Tompson.
\newblock Addressing sample inefficiency and reward bias in inverse
  reinforcement learning.
\newblock {\em CoRR}, abs/1809.02925, 2018.

\bibitem{hausman2018multi}
Karol Hausman, Yevgen Chebotar, Stefan Schaal, Gaurav Sukhatme, and Joseph Lim.
\newblock Multi-modal imitation learning from unstructured demonstrations using
  generative adversarial nets.
\newblock In {\em 31st Annual Conference on Neural Information Processing
  Systems (NIPS 2017)}, pages 1236--1246. Curran Associates, Inc., 2018.

\bibitem{sutton1999between}
Richard~S Sutton, Doina Precup, and Satinder Singh.
\newblock Between mdps and semi-mdps: A framework for temporal abstraction in
  reinforcement learning.
\newblock {\em Artificial intelligence}, 112(1-2):181--211, 1999.

\bibitem{kulkarni2016hierarchical}
Tejas~D Kulkarni, Karthik Narasimhan, Ardavan Saeedi, and Josh Tenenbaum.
\newblock Hierarchical deep reinforcement learning: Integrating temporal
  abstraction and intrinsic motivation.
\newblock {\em Advances in neural information processing systems},
  29:3675--3683, 2016.

\bibitem{nachum2018data}
Ofir Nachum, Shixiang~Shane Gu, Honglak Lee, and Sergey Levine.
\newblock Data-efficient hierarchical reinforcement learning.
\newblock {\em Advances in Neural Information Processing Systems},
  31:3303--3313, 2018.

\bibitem{nachum2019does}
Ofir Nachum, Haoran Tang, Xingyu Lu, Shixiang Gu, Honglak Lee, and Sergey
  Levine.
\newblock Why does hierarchy (sometimes) work so well in reinforcement
  learning?
\newblock {\em arXiv preprint arXiv:1909.10618}, 2019.

\bibitem{henderson2018optiongan}
Peter Henderson, Wei-Di Chang, Pierre-Luc Bacon, David Meger, Joelle Pineau,
  and Doina Precup.
\newblock Optiongan: Learning joint reward-policy options using generative
  adversarial inverse reinforcement learning.
\newblock In {\em Proceedings of the AAAI conference on artificial
  intelligence}, volume~32, 2018.

\bibitem{sharma2018directed}
Mohit Sharma, Arjun Sharma, Nicholas Rhinehart, and Kris~M Kitani.
\newblock Directed-info gail: Learning hierarchical policies from unsegmented
  demonstrations using directed information.
\newblock In {\em International Conference on Learning Representations}, 2018.

\bibitem{jing2021adversarial}
Mingxuan Jing, Wenbing Huang, Fuchun Sun, Xiaojian Ma, Tao Kong, Chuang Gan,
  and Lei Li.
\newblock Adversarial option-aware hierarchical imitation learning.
\newblock {\em arXiv preprint arXiv:2106.05530}, 2021.

\bibitem{codevilla2018end}
Felipe Codevilla, Matthias M{\"u}ller, Antonio L{\'o}pez, Vladlen Koltun, and
  Alexey Dosovitskiy.
\newblock End-to-end driving via conditional imitation learning.
\newblock In {\em 2018 IEEE International Conference on Robotics and Automation
  (ICRA)}, pages 4693--4700. IEEE, 2018.

\bibitem{hertweck2020simple}
Tim Hertweck, Martin Riedmiller, Michael Bloesch, Jost~Tobias Springenberg,
  Noah Siegel, Markus Wulfmeier, Roland Hafner, and Nicolas Heess.
\newblock Simple sensor intentions for exploration.
\newblock {\em arXiv preprint arXiv:2005.07541}, 2020.

\bibitem{haarnoja2018soft}
Tuomas Haarnoja, Aurick Zhou, Pieter Abbeel, and Sergey Levine.
\newblock Soft actor-critic: Off-policy maximum entropy deep reinforcement
  learning with a stochastic actor.
\newblock In {\em ICML}, 2018.

\bibitem{cabi2017intentional}
Serkan Cabi, Sergio~G{\'o}mez Colmenarejo, Matthew~W Hoffman, Misha Denil, Ziyu
  Wang, and Nando Freitas.
\newblock The intentional unintentional agent: Learning to solve many
  continuous control tasks simultaneously.
\newblock In {\em Conference on Robot Learning}, pages 207--216. PMLR, 2017.

\bibitem{jaderberg2016reinforcement}
Max Jaderberg, Volodymyr Mnih, Wojciech~Marian Czarnecki, Tom Schaul, Joel~Z
  Leibo, David Silver, and Koray Kavukcuoglu.
\newblock Reinforcement learning with unsupervised auxiliary tasks.
\newblock {\em arXiv preprint arXiv:1611.05397}, 2016.

\bibitem{colgateCobotsRobotsCollaboration1996}
J.~Edward Colgate, J.~Edward, Michael~A. Peshkin, and Witaya Wannasuphoprasit.
\newblock Cobots: {{Robots For Collaboration With Human Operators}}, 1996.

\bibitem{kipf2019compositional}
Thomas Kipf, Yujia Li, Hanjun Dai, Vinicius Zambaldi, Alvaro Sanchez-Gonzalez,
  Edward Grefenstette, Pushmeet Kohli, and Peter Battaglia.
\newblock Compile: Compositional imitation learning and execution.
\newblock In {\em International Conference on Machine Learning (ICML)}, 2019.

\bibitem{lee2019composing}
Youngwoon Lee, Shao-Hua Sun, Sriram Somasundaram, Edward~S. Hu, and Joseph~J.
  Lim.
\newblock Composing complex skills by learning transition policies.
\newblock In {\em Proceedings of International Conference on Learning
  Representations}, 2019.

\bibitem{aroraSurveyInverseReinforcement2018}
Saurabh Arora and Prashant Doshi.
\newblock A {{Survey}} of {{Inverse Reinforcement Learning}}: {{Challenges}},
  {{Methods}} and {{Progress}}.
\newblock {\em arXiv:1806.06877 [cs, stat]}, June 2018.

\bibitem{rl_sandbox}
Bryan Chan.
\newblock Rl sandbox.
\newblock \url{https://github.com/chanb/rl_sandbox_public}, 2020.

\bibitem{paszke2019pytorch}
Adam Paszke, Sam Gross, Francisco Massa, Adam Lerer, James Bradbury, Gregory
  Chanan, Trevor Killeen, Zeming Lin, Natalia Gimelshein, Luca Antiga, et~al.
\newblock Pytorch: An imperative style, high-performance deep learning library.
\newblock {\em Advances in neural information processing systems},
  32:8026--8037, 2019.

\bibitem{gulrajani2017improved}
Ishaan Gulrajani, Faruk Ahmed, Martin Arjovsky, Vincent Dumoulin, and Aaron
  Courville.
\newblock Improved training of wasserstein gans.
\newblock {\em arXiv preprint arXiv:1704.00028}, 2017.

\bibitem{kingma2013auto}
Diederik~P Kingma and Max Welling.
\newblock Auto-encoding variational bayes.
\newblock {\em arXiv preprint arXiv:1312.6114}, 2013.

\bibitem{fujimoto2018addressing}
Scott Fujimoto, Herke Hoof, and David Meger.
\newblock Addressing function approximation error in actor-critic methods.
\newblock In {\em International Conference on Machine Learning}, pages
  1587--1596. PMLR, 2018.

\bibitem{van2016deep}
Hado Van~Hasselt, Arthur Guez, and David Silver.
\newblock Deep reinforcement learning with double q-learning.
\newblock In {\em Proceedings of the AAAI conference on artificial
  intelligence}, volume~30, 2016.

\bibitem{mnih2013playing}
Volodymyr Mnih, Koray Kavukcuoglu, David Silver, Alex Graves, Ioannis
  Antonoglou, Daan Wierstra, and Martin Riedmiller.
\newblock Playing atari with deep reinforcement learning.
\newblock {\em arXiv preprint arXiv:1312.5602}, 2013.

\bibitem{haarnoja2018soft2}
Tuomas Haarnoja, Aurick Zhou, Kristian Hartikainen, George Tucker, Sehoon Ha,
  Jie Tan, Vikash Kumar, Henry Zhu, Abhishek Gupta, Pieter Abbeel, et~al.
\newblock Soft actor-critic algorithms and applications.
\newblock {\em arXiv preprint arXiv:1812.05905}, 2018.

\bibitem{coumans2016pybullet}
Erwin Coumans and Yunfei Bai.
\newblock Pybullet, a python module for physics simulation for games, robotics
  and machine learning.
\newblock 2016.

\bibitem{kingma2014adam}
Diederik~P Kingma and Jimmy Ba.
\newblock Adam: A method for stochastic optimization.
\newblock {\em arXiv preprint arXiv:1412.6980}, 2014.

\end{thebibliography}

\newpage
\appendix

\section{Learning from Guided Play Algorithm}
The complete pseudo-code is given in Algorithm \ref{alg:DACX}. Our implementation builds on RL Sandbox \cite{rl_sandbox}, an open-source PyTorch \cite{paszke2019pytorch} implementation for RL algorithms. For learning the discriminators, we apply gradient penalty to regularize the discriminators \cite{gulrajani2017improved}, as done in DAC \cite{DBLP:journals/corr/abs-1809-02925}. We optimize the intentions via the reparameterization trick \cite{kingma2013auto}. As commonly done in deep RL algorithms, we use the Clipped Double Q-Learning trick \cite{fujimoto2018addressing} to mitigate overestimation bias \cite{van2016deep} and use a target network to mitigate learning instability \cite{mnih2013playing} when training the Q-functions. We also learn the temperature parameter $\alpha_{\tasks}$ separately for each task $\tasks$ (see Section 5 of \cite{haarnoja2018soft2} for more details on learning $\alpha$). The hyperparameters are provided in \Cref{sec:hyperparameters}.
Please see attached video for a short representative example of what LfGP looks like in practice.

\begin{algorithm}[htb]
\caption{Learning from Guided Play (LfGP)}
\label{alg:DACX}
\textbf{Input}: Expert replay buffers $\buffer^E_{\text{main}}, \buffer^E_{1}, \dots, \buffer^E_{K}$, scheduler period $\xi$, sample batch size $N$\\
\textbf{Parameters}: Intentions $\pi_\tasks$ with corresponding Q-functions $Q_\tasks$ and discriminators $D_\tasks$, and scheduler $\pi_S$ (e.g. with Q-table $Q_S$)
\begin{algorithmic}[1] %
\STATE Initialize replay buffer $\mathcal{B}$ \\
\FOR{$t = 1, \dots,$}
    \STATE \# Interact with environment
    \STATE For every $\xi$ steps, select intention $\pi_\tasks$ using $\pi_S$
    \STATE Select action $a_t$ using $\pi_\tasks$
    \STATE Execute action $a_t$ and observe next state $s'_t$
    \STATE Store transition $\langle s_t, a_t, s'_t \rangle$ in $\mathcal{B}$
    \STATE 
    \STATE \# Update discriminator $D_{\tasks'}$ for each task $\tasks'$
    \STATE Sample $\left\{ (s_i, a_i) \right\}_{i=1}^{N} \sim \mathcal{B}$
    \FOR{each task $\tasks'$}
        \STATE Sample $\left\{ (s'_i, a'_i) \right\}_{i=1}^{B} \sim \buffer^E_k$
        \STATE Update $D_{\tasks'}$ following equation 3 using GAN + Gradient Penalty
    \ENDFOR
    \STATE
    \STATE \# Update intentions $\pi_{\tasks'}$ and Q-functions $Q_{\tasks'}$ for each task $\tasks'$
    \STATE Sample $\left\{ (s_i, a_i) \right\}_{i=1}^{N} \sim \mathcal{B}$
    \STATE Compute reward $D_{\tasks'}(s_i, a_i)$ for each task $\tasks'$
    \STATE Update $\pi$ and $Q$ following equations 7 and 8
    \STATE
    \STATE \# Update scheduler $\pi_S$ if necessary
    \IF{at the end of effective horizon}
        \STATE Compute main task return $G_{\tasks_\text{main}}$ using reward estimate from $D_\text{main}$
        \STATE Update $\pi_S$ (e.g. update Q-table $Q_S$ following equation 12 and recompute Boltzmann distribution)
    \ENDIF
\ENDFOR
\end{algorithmic}
\vspace{2mm}
\end{algorithm}

\section{Environment Details} \label{sec:env_details}
\begin{wrapfigure}{R}{.5\textwidth}
	\centering
	\includegraphics[width=0.48\textwidth]{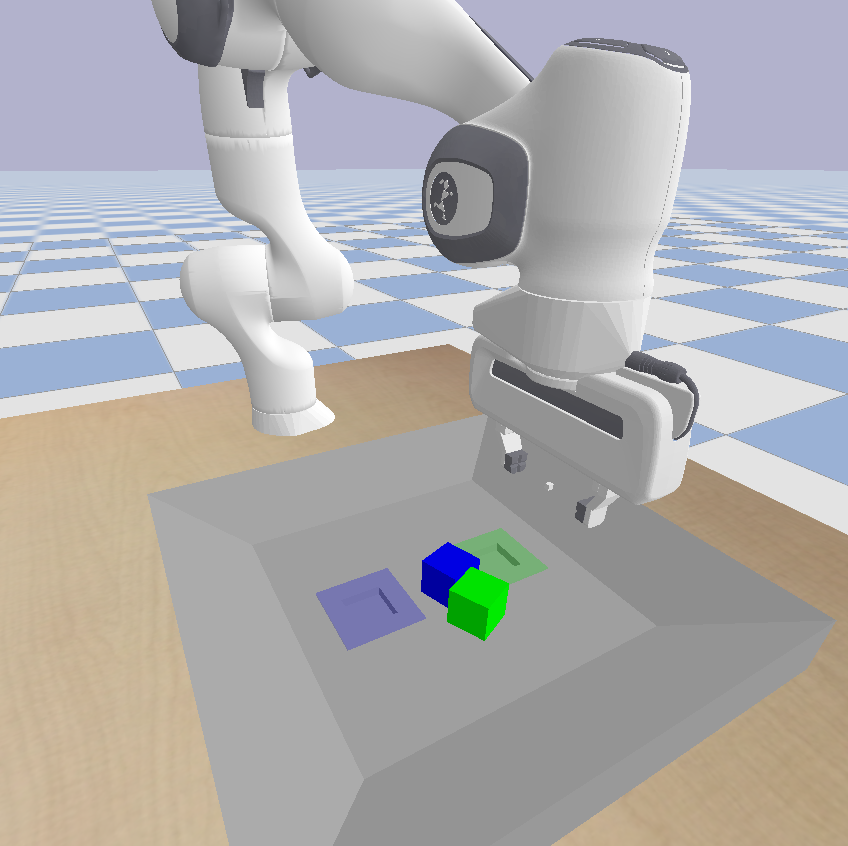}
	\caption{An image of our multitask environment immediately after a reset.}
	\label{fig:env_example}
\end{wrapfigure}
\begin{table}
	\centering
	\small
    \caption{The components used in our environment observations, common to all tasks. Grip finger position is a continuous value from 0 (closed) to 1 (open).}
	\begin{tabular}{lrlll}
	    \toprule
		    Component  & Dim            & Unit       & Privileged?  &  Extra info                        \\
		\midrule
		EE pos.                & 3              & m          & No           & rel. to base           \\
		EE velocity            & 3              & m/s        & No           & rel. to base           \\
		Grip finger pos.       & 6              & [0, 1]       & No           & current, last 2        \\
		Block pos.            & 6              & m          & Yes            & both blocks           \\
		Block rot.            & 8              & quat       & Yes            & both blocks           \\
		Block trans vel.      & 6              & m/s        & Yes            & rel. to base           \\
		Block rot vel.        & 6              & rad/s      & Yes            & rel. to base           \\
		Block rel to EE       & 6              & m          & Yes            & both blocks           \\
		Block rel to block    & 3              & m          & Yes            & in base frame           \\
		Block rel to slot     & 6              & m          & Yes            & both blocks           \\
		Force-torque          & 6              & N,Nm       & No            & at wrist           \\
		\midrule
		\textbf{Total}        & \textbf{59}      &       &             &         \\
	\end{tabular}
	\label{tab:obs_details}
\end{table}
A screenshot of our environment, simulated in PyBullet \cite{coumans2016pybullet}, is shown in \Cref{fig:env_example}.
We chose this environment because we desired tasks that a) have a large distribution of possible initial states, representative of manipulation tasks in the real world, b) have a shared observation/action space with several other tasks, allowing the use of auxiliary tasks and transfer learning, and c) require a reasonably long horizon and significant use of contact to solve.
The environment contains a tray with sloped edges to keep the blocks within the reachable workspace of the end-effector, as well as a green and a blue block, each of which are 4 cm $\times$ 4 cm $\times$ 4 cm and set to a mass of 100 g.
The dimensions of the lower part of the tray, before reaching the sloped edges, are 30 cm $\times$ 30 cm.
The dimensions of the bring boundaries (shaded blue and green regions) are 8 cm $\times$ 8 cm, while the dimensions of the insertion slots, which are directly in the center of each shaded region, are 4.1 cm $\times$ 4.1 cm $\times$ 1 cm.
The boundaries for end-effector movement, relative to the tool center point that is directly between the gripper fingers, are a 30 cm $\times$ 30 cm $\times$ 14.5 cm box, where the bottom boundary is low enough to allow the gripper to interact with objects, but not to collide with the bottom of the tray.

See \Cref{tab:obs_details} for a summary of our environment observations.
In this work, we use privileged state information (e.g., block poses), but adapting our method to exclusively use image-based data is straightforward since we do not use hand-crafted reward functions as in \cite{riedmiller2018learning}.

The environment movement actions are 3-DOF translational position changes, where the position change is relative to the current end-effector position, and we leverage PyBullet's built-in position-based inverse kinematics function to generate joint commands.
Our actions also contain a fourth dimension for actuating the gripper.
To allow for the use of policy models with exclusively continuous outputs, this dimension accepts any real number, with any value greater than 0 commanding the gripper to open, and any number lower than 0 commanding it to close.
Actions are supplied at a rate of 20 Hz, and each training episode is limited to being 18 seconds long, corresponding to 360 time steps per episode.
For play-based expert data collection, we also reset the environment manually every 360 time steps.
Between episodes, block positions are randomized to any pose within the tray, and the end-effector is randomized to any position between 5 and 14.5 cm above the tray, within the earlier stated end-effector bounds, with the gripper fully opened.
The only exception to these initial conditions is during expert data collection and agent training of the Unstack-Stack task: in this case, the green block is manually set to be on top of the blue block at the start of the episode.

\section{Procedure for Obtaining Experts} \label{sec:expert_gen}
As stated, we used SAC-X \cite{riedmiller2018learning} to train models that we used for generating expert data.
We used the same hyperparameters as we used for LfGP (see \Cref{tab:hyperparameters_ail}), apart from the discriminator which, of course, does not exist in SAC-X.
See \Cref{sec:evaluation} for details on the hand-crafted rewards that we used for training these models.
For an example of gathering play-based expert data, please see our attached video.

We made two modifications to regular SAC-X to speed up learning.
First, we pre-trained a Move-Object model before transferring it to each of our main tasks, as we did in Section 5.3 of our main paper, since we found that SAC-X would plateau when we tried to learn the more challenging tasks from scratch.
The need for this modification demonstrates another noteworthy benefit of LfGP---when training LfGP, main tasks could be learned from scratch, and generally in fewer time steps, than it took to train our experts.
Second, during the transfer to the main tasks, we used what we called a conditional weighted scheduler instead of a Q-Table: we defined weights for every combination of tasks, so that the scheduler would pick each task with probability $P(\tasks^{(h)} | \tasks^{(h-1)})$, ensuring that $\forall \tasks' \in \tasks_{\text{all}}, \sum_{\tasks \in \tasks_{\text{all}}} P(\tasks | \tasks') = 1$.
The weights that we used were fairly consistent between main tasks, and can be found in our included code.
The conditional weighed scheduler ensured that every task was still explored throughout the learning process, ensuring that we would have high-quality experts for every auxiliary task, in addition to the main task.

\section{Evaluation} \label{sec:evaluation}
As stated in our paper, we evaluated all algorithms by testing the mean output of the main-task policy head in our environment and generating a success rate based on 50 randomly selected resets.
These evaluation episodes were all run for 360 time steps to match our training environment, and if a condition for success was met within that time, they were recorded as a success. 
See our included video for sample runs.
The remaining section describes in detail how we evaluated success for each of our main and auxiliary tasks.

As previously stated, we also trained experts using modified SAC-X \cite{riedmiller2018learning} that required us to define a set of reward functions for each task as well, which we also include in this section.
The authors of \cite{riedmiller2018learning} focused on sparse rewards, but also showed a few experiments in which dense rewards reduced the time to learn adequate policies, so we also used dense rewards.
We would like to note that many of these reward functions are particularly complex and required significant manual shaping effort, further motivating the use of an imitation learning scheme like the one presented in this paper.
It is possible that we could have gotten away with sparse rewards, such as those used in \cite{riedmiller2018learning}, but our compute resources made this impractical---for example, in \cite{riedmiller2018learning}, their agent took 5000 episodes $\times$ 36 actors $\times$ 360 time steps $=$ 64.8 M time steps to learn their stacking task, which would have taken over a month of wall-time on our fastest machine.
To see the specific values used for the rewards and success conditions described in these sections, see our included code.

Unless otherwise stated, each of the success conditions in this section had to be held for 10 time steps, or 0.5 seconds, before they registered as a success.
This was to prevent registering a success when, for example, the blue block slipped off the green block during Stack.

\subsection{Common}
For each of these functions, we use the following common labels:
\begin{itemize}
    \item $p_b$: blue block position,
    \item $v_b$: blue block velocity,
    \item $a_b$: blue block acceleration,
    \item $p_g$: green block position,
    \item $p_e$: end-effector tool center point position (TCP),
    \item $p_s$: center of a block pushed into one of the slots,
    \item $g_1$: (scalar) gripper finger 1 position,
    \item $g_2$: (scalar) gripper finger 2 position, and
    \item $a_g$: (scalar) gripper open/close action.
\end{itemize}
A block is flat on the tray when $p_{b,z} = 0$ or $p_{g,z} = 0$.
To further reduce training time for SAC-X experts, all rewards were set to 0 if $\lVert p_b - p_e \rVert > 0.1$ and $\lVert p_g - p_e \rVert > 0.1$ (i.e., the TCP must be within 10 cm of either block).
During training while using the Unstack-Stack variation of our environment, a penalty of -0.1 was added to each reward if $\lVert p_{g,z} \rVert > 0.001$ (i.e., there was a penalty to all rewards if the green block was not flat on the tray).

\subsection{Stack/Unstack-Stack}
The evaluation conditions for Stack and Unstack-Stack are identical, but in our Unstack-Stack experiments, the environment is manually set to have the green block start on top of the blue block.

\subsubsection{Success}
Using internal PyBullet commands, we check to see whether the blue block is in contact with the green block and is \textit{not} in contact with both the tray and the gripper.

\subsubsection{Reward}
We include a term for checking the distance between the blue block and the spot above the the green block, a term for rewarding increasing distance between the block and the TCP once the block is stacked, a term for shaping lifting behaviour, a term for rewarding closing the gripper when the block is within a tight reaching tolerance, and a term for rewarding the opening the gripper once the block is stacked.

\subsection{Bring/Insert}
We use the same success and reward calculations for Bring and Insert, but for Bring the threshold for success is 3 cm, and for insert, it is 2.5 mm.

\subsubsection{Success}
We check that the distance between $p_b$ and $p_s$ is less than the defined threshold, that the blue block is touching the tray, and that the end-effector is \textit{not} touching the block.
For insert, the block can only be within 2.5 mm of the insertion target if it is correctly inserted.

\subsubsection{Reward}
We include a term for checking the distance between the $p_b$ and $p_s$ and a term for rewarding increasing distance between $p_b$ and $p_e$ once the blue block is brought/inserted.

\subsection{Open-Gripper/Close-Gripper}
We use the same success and reward calculations for Open-Gripper and Close-Gripper, apart from inverting the condition.

\subsubsection{Success}
For Open-Gripper and Close-Gripper, we check to see if $a_g < 0$ or $a_g > 0$ respectively.

\subsubsection{Reward}
We include a term for checking the action, as we do in the success condition, and also include a shaping term that discourages high magnitudes of the movement action.

\subsection{Lift}
\subsubsection{Success}
We check to see if $p_{b,z} > 0.06$.

\subsubsection{Reward}
We add a dense reward for checking the height of the block, but specifically also check that the gripper positions correspond to being closed around the block, so that the block does not simply get pushed up the edges of the tray.
We also include a shaping term for encouraging the gripper to close when the block is reached.

\subsection{Reach}
\subsubsection{Success}
We check to see if $\lVert p_e - p_b \rVert < 0.015$.

\subsubsection{Reward}
We have a single dense term to check the distance between $p_e$ and $p_b$.

\subsection{Move-Object}
For Move-Object, we changed the required holding time for success to 1 second, or 20 time steps.

\subsubsection{Success}
We check to see if the $v_b > 0.05$ and $a_b < 5$.
The acceleration condition ensures that the arm has learned to move the block in smooth trajectories, rather than vigorously shaking it or continuosly picking up and dropping it.

\subsubsection{Reward}
We include a velocity term and an acceleration penalty, as in the success condition, but also include a dense bonus for lifting the block.

\section{Return Plots}
\begin{figure*}[ht]
	\centering
	\includegraphics[width=.95\textwidth]{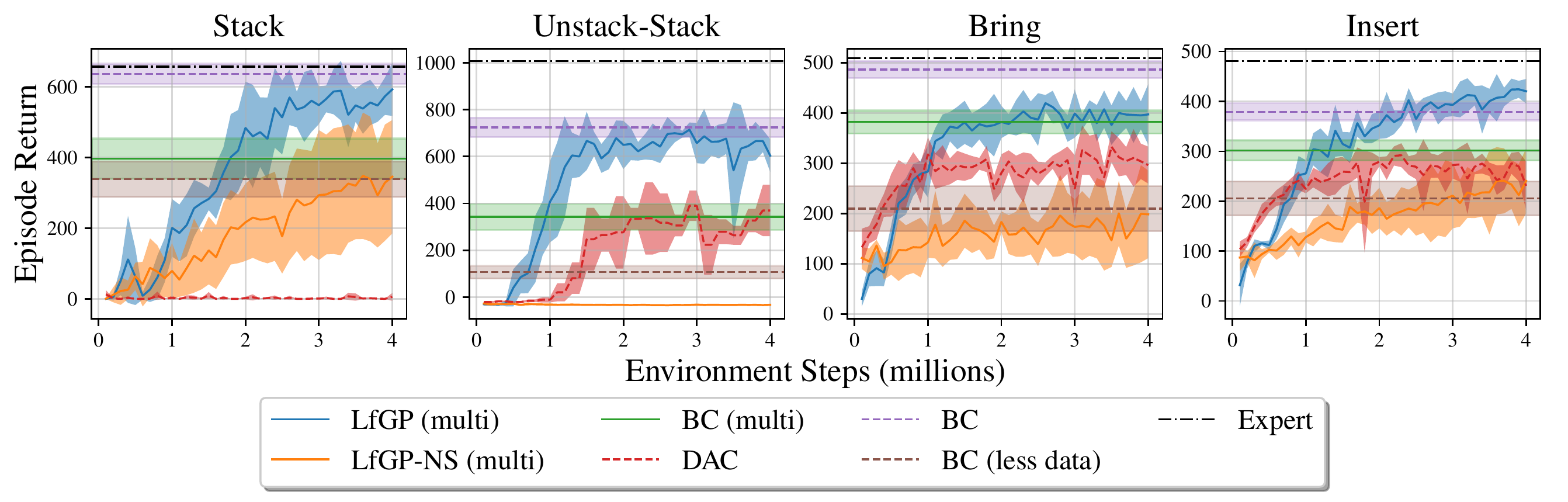}
	\caption{Episode return for LfGP compared with all baselines. Shaded area corresponds to standard deviation.}
	\label{fig:main_return}
\end{figure*}
\begin{figure*}[ht]
	\centering
	\includegraphics[width=.95\textwidth]{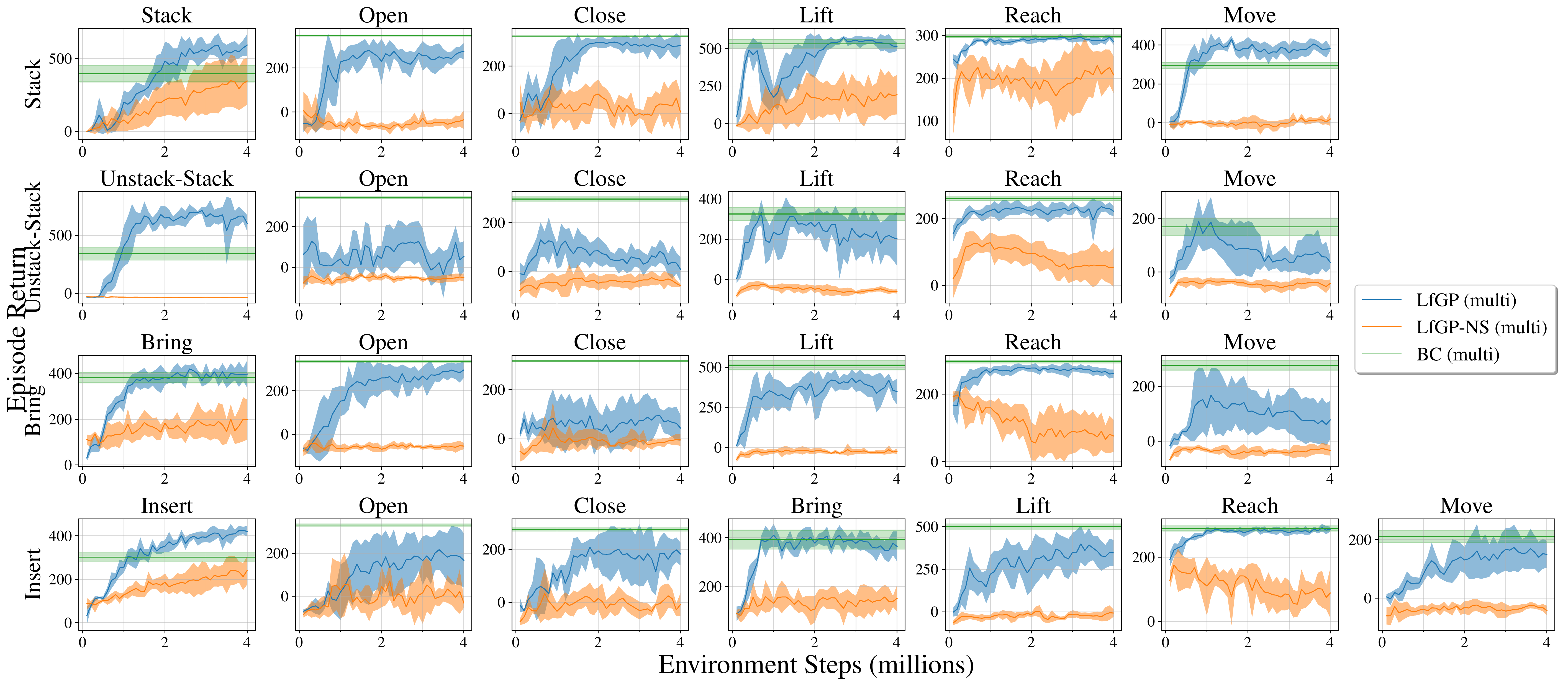}
	\caption{Episode return for LfGP compared with multitask baselines on all tasks. Shaded area corresponds to standard deviation.}
	\label{fig:multitask_return}
\end{figure*}

\begin{figure}[t]
	\centering
	\begin{subfigure}{0.45\textwidth}
	    \includegraphics[width=\textwidth]{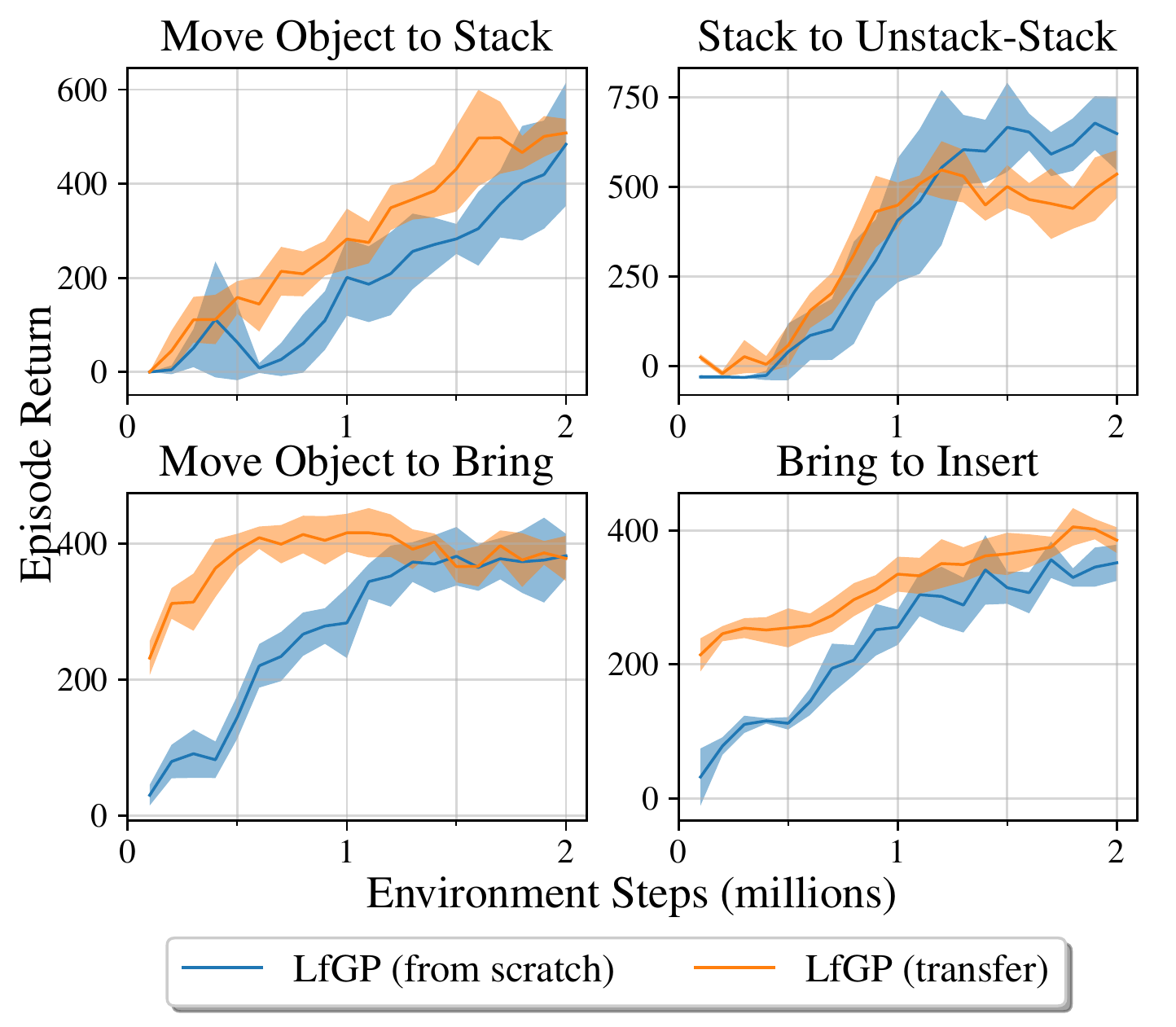}
	\end{subfigure}
	\hfill
	\begin{subfigure}{0.45\textwidth}
	    \includegraphics[width=\textwidth]{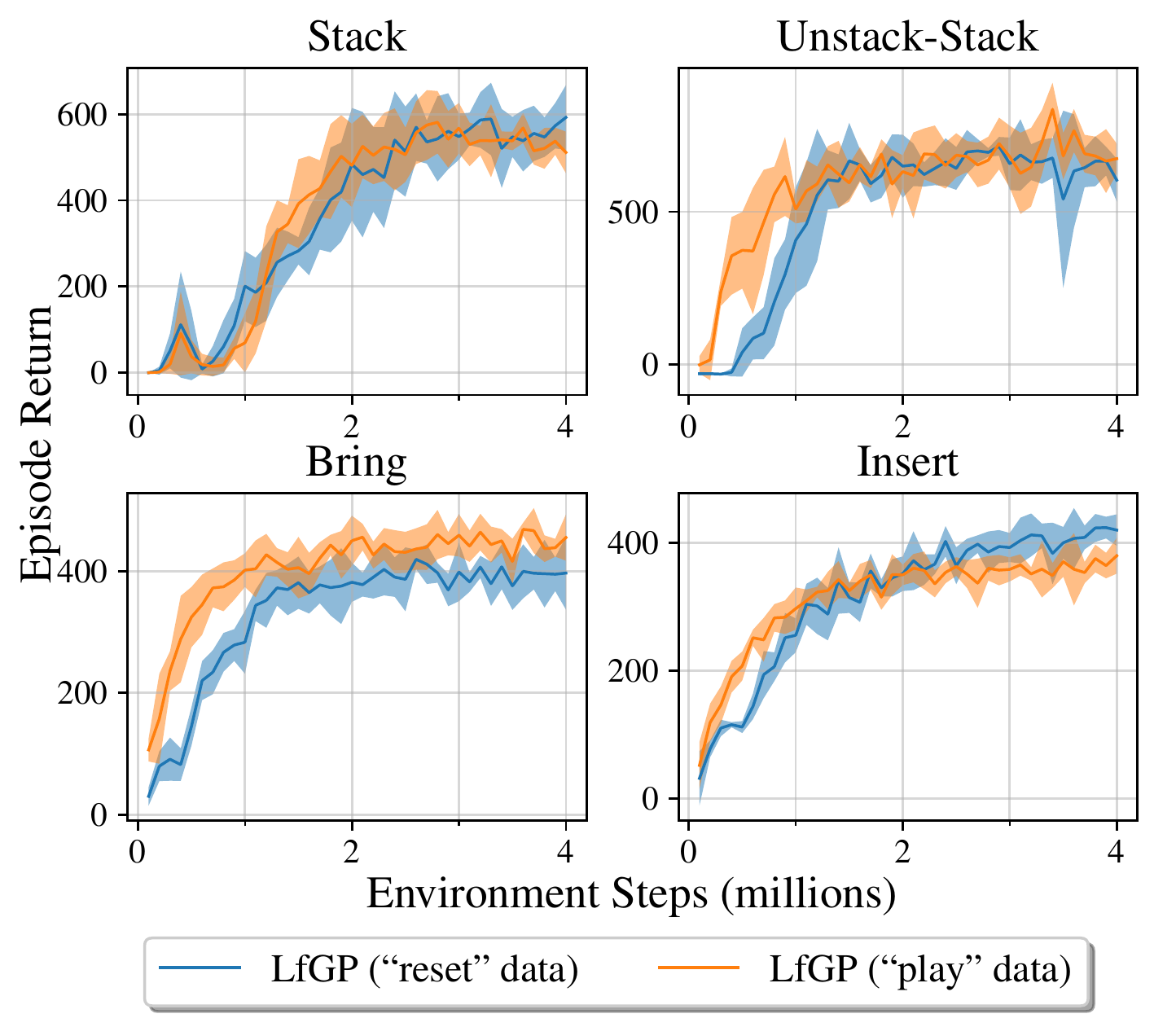}
	\end{subfigure}
	\caption{\textbf{Left:} Episode return for our transfer experiments. \textbf{Right:} Episode return for our play-based expert data experiments. Shaded area corresponds to standard deviation. }
	\label{fig:transfer_and_play_return}
\end{figure}

As previously stated, we generated hand-crafted reward functions for each of our tasks for the purpose of training our SAC-X experts.
Given that we have these rewards, we can also generate return plots corresponding to our results to add extra insight.
The episode return plots corresponding to our main task performance (Figure 3 of the main paper), multitask performance (Figure 4 of the main paper), transfer performance (Figure 5 of the main paper) and play-based expert data performance (Figure 6 of the main paper) are shown in \Cref{fig:main_return}, \Cref{fig:multitask_return}, and \Cref{fig:transfer_and_play_return} respectively.
The patterns displayed in these plots are, for the most part, quite similar to the success rate plots.
One notable exception was the fact that in Unstack-Stack, DAC performed far worse than LfGP as measured by return, as opposed to success rate---this can be explained by the fact that the DAC policies learned to unstack and restack the blue block continually, rather than letting the blue block rest on top of the green block (see included videos).
As well, in the transfer experiments, it becomes clear that transferring from existing models did, in fact, have a notable increase in training speed for \textit{all} tasks, which was not necessarily as evident from observing the success rate plots.

\section{Model Architectures and Hyperparameters} \label{sec:hyperparameters}
All the single-task models share the same network architectures and all the multitask models share the same network architectures. All layers are initialized using the PyTorch default methods \cite{paszke2019pytorch}.

For the single-task variant, the policy is a fully-connected network with two hidden layers followed by ReLU activation. 
Each hidden layer consists of 256 hidden units. 
The output of the policy is split into two vectors, mean $\hat{\mu}$ and variance $\hat{\sigma}^2$.  
The vectors are used to construct a Gaussian distribution (i.e. $N(\hat{\mu}, \hat{\sigma}^2 \mathbf{I})$, where $\mathbf{I}$ is the identity matrix). When computing actions, we squash the samples using the tanh function, and bounding the actions to be in range $[-1, 1]$, as done in SAC \cite{haarnoja2018soft2}.
The variance $\hat{\sigma}^2$ is computed by applying a softplus function followed by a sum with an epsilon $\epsilon =$ 1e-7 to prevent underflow: $\hat{\sigma}_i = \text{softplus}(\hat{x}_i) + \epsilon$.
The Q-functions are fully-connected networks with two hidden layers followed by ReLU activation.
Each hidden layer consists of 256 hidden units.
The output of the Q-function is a scalar corresponding to the value estimate given the current state-action pair.
Finally, The discriminator is a fully-connected network with two hidden layers followed by tanh activation.
Each hidden layer consists of 256 hidden units.
The output of the discriminator is a scalar corresponding to the logits to the sigmoid function.
The sigmoid function can be viewed as the probability of the current state-action pair coming from the expert distribution.

For multitask variant, the policies and the Q-functions share their initial layers.
There are two shared fully-connected layers followed by ReLU activation.
Each layer consists of 256 hidden units.
The output of the last shared layer is then fed into the policies and Q-functions.
Each policy head and Q-function head correspond to one task and have the same architecture: a two-layered fully-connected network followed by ReLU activations.
The output of the policy head corresponds to the parameters of a Gaussian distribution, as described previously.
Similarly, the output of the Q-function head corresponds to the value estimate.
Finally, The discriminator is a fully-connected network with two hidden layers followed by tanh activation.
Each hidden layer consists of 256 hidden units.
The output of the discriminator is a vector, where the $i^\text{th}$ entry corresponds to the logit to the sigmoid function for task $\tasks_i$.
The $i^\text{th}$ sigmoid function corresponds to the probability of the current state-action pair coming from the expert distribution in task $\tasks_i$.

The hyperparameters for our experiments are listed in \Cref{tab:hyperparameters_ail} and \Cref{tab:hyperparameters_bc}.
In BC, \textit{overfit tolerance} refers to the number of full dataset training epochs without an improvement in validation error before we stop training.
All models are optimized using Adam Optimizer \cite{kingma2014adam} with PyTorch default values, unless specified otherwise.

\begin{table}[ht]
	\centering
	\small
	\caption{Hyperparameters for AIL algorithms across all tasks.}
	\begin{tabularx}{.7\columnwidth}{llll}
	    \toprule
        Algorithm & LfGP (Ours) & LfGP-NS & DAC\\
        \midrule
        \midrule
        Total Interactions & \multicolumn{3}{c}{4M} \\
        Buffer Size & \multicolumn{3}{c}{4M} \\
		Buffer Warmup & \multicolumn{3}{c}{1000} \\
		Initial Exploration & \multicolumn{3}{c}{1000} \\
        \midrule
        \midrule
		\textit{Intention} & & & \\
		$\gamma$ &  \multicolumn{3}{c}{0.99}  \\
		Batch Size & \multicolumn{3}{c}{256}  \\
		$Q$ Update Freq. & \multicolumn{3}{c}{1}  \\
		Target $Q$ Update Freq. & \multicolumn{3}{c}{1} \\
		$\pi$ Update Freq. & \multicolumn{3}{c}{1} \\
		Polyak Averaging & \multicolumn{3}{c}{0.005} \\
        $Q$ Learning Rate & \multicolumn{3}{c}{3e-4} \\
		$\pi$ Learning Rate & \multicolumn{3}{c}{1e-5} \\
        $\alpha$ Learning Rate & \multicolumn{3}{c}{3e-4} \\
        Initial $\alpha$ & \multicolumn{3}{c}{1} \\
        Target Entropy & \multicolumn{3}{c}{4} \\
        Max. Gradient Norm & \multicolumn{3}{c}{10} \\
        \midrule
        \midrule
		\textit{Discriminator} & & & \\
		Learning Rate & \multicolumn{3}{c}{3e-4} \\
        Batch Size & \multicolumn{3}{c}{256} \\
        Gradient Penalty $\lambda$ & \multicolumn{3}{c}{10} \\
        \midrule
        \midrule
		\textit{Scheduler} & & & \\
		Type & Q-table & Select $\tasks_\text{main}$ & N/A \\
		$\xi$ & 45 & N/A & N/A \\
		$\phi$ & 0.6 & N/A & N/A \\
		Initial Temp. & 360 & N/A & N/A \\
		Temp. Decay & 0.9995 & N/A & N/A \\
		Min. Temp. & 0.1 & N/A & N/A \\
		\bottomrule
	\end{tabularx}
	\label{tab:hyperparameters_ail}
\end{table}

\begin{table}[ht]
	\centering
	\small
	\caption{Hyperparameters for BC algorithms across all tasks.}
	\begin{tabularx}{0.7\columnwidth}{c|ccc}
	    \toprule
        Algorithm & BC & BC (Less Data) & Multitask BC \\
        \midrule
        \midrule
		Batch Size & \multicolumn{3}{c}{256} \\
        Learning Rate & \multicolumn{3}{c}{3e-4} \\
        Overfit Tolerance & \multicolumn{3}{c}{100} \\
		\bottomrule
	\end{tabularx}
	\label{tab:hyperparameters_bc}
\end{table}

\section{Experimental Hardware}
For a list of the software we used in this work, see our included code and instructions.
We used a number of different computers for completing experiments:
\begin{enumerate}
    \item GPU: NVidia Quadro RTX 8000, CPU: AMD - Ryzen 5950x 3.4 GHz 16-core 32-thread, RAM: 64GB, OS: Ubuntu 20.04.
    \item GPU: NVidia V100 SXM2, CPU: Intel Gold 6148 Skylake @ 2.4 GHz (only used 4 threads), RAM: 32GB, OS: CentOS 7.
    \item GPU: Nvidia GeForce RTX 2070, CPU: RYZEN Threadripper 2990WX, RAM: 32GB, OS: Ubuntu 20.04.
\end{enumerate}

\section{Open-Action and Close-Action Distribution Matching}
There was one exception to the ``reset-based'' method we used for collecting our expert data.
Specifically, our Open-Gripper and Close-Gripper tasks required several additional considerations.
It is worth reminding the reader that our Open-Gripper and Close-Gripper tasks were meant to simply open or close the gripper, respectively, while remaining reasonably close to either block.
If we were to use the approach described above verbatim, the Open-Gripper and Close-Gripper data would contain no $(s,a)$ pairs where the gripper actually released or grasped the block, instead immediately opening or closing the gripper and simply hovering near the blocks.
Perhaps unsurprisingly, this was detrimental to our algorithm's performance: as one example, an agent attempting to learn Stack would, if Open-Gripper was selected while the blue block was held above the green block, move the currently grasped blue block \textit{away} from the green block before dropping it on the tray.
This behaviour, of course, is not what we would want, but it better matches an expert distribution collected using the method described above.

To mitigate this, our Open-Gripper data actually contain a mix of each of the other sub-tasks called first for 45 time steps, followed by a switch to Open-Gripper, ensuring that the expert dataset contains some degree of block-releasing, with the trade-off being that 25\% of the Open-Gripper expert data is specific to whatever the main task is.
We left this detail out of our main paper for clarity, since it corresponds to only 4-5\% of the data ($2250 / 45000$ or $2250 / 54000$) that was claimed as being reusable being, in actuality, task-specific.
Similarly, the Close-Gripper data calls Lift for 15 time steps before switching to Close-Gripper, ensuring that the Close-gripper dataset will contain a large proportion of data where the block is actually grasped.
Given the simplicity of designing a reward function in these two cases, a natural question is whether Open-Gripper and Close-Gripper could use hand-crafted reward functions, or even hand-crafted policies, instead of these specialized datasets.
In our experiments, both of these alternatives proved to be quite detrimental to our algorithm, so we leave further exploration of these options for future work.

\end{document}